%% file: main.tex
\documentclass[10pt,twocolumn,letterpaper]{article}
\usepackage{thunlp_cvpr}

\usepackage{graphicx}
\usepackage{booktabs}

\usepackage{microtype}
\usepackage{booktabs}      
\usepackage{multirow}       
\usepackage{graphicx}       
\usepackage[table]{xcolor}
\usepackage{inconsolata}
\usepackage{graphicx}
\usepackage{amsmath}
\usepackage{booktabs}
\usepackage{array}
\usepackage{enumitem}
\usepackage{caption}
\usepackage{tabularx}
\usepackage{float}
\usepackage{dblfloatfix}

\usepackage{eso-pic} 

\newcommand{\addtitlesideimage}[1]{%
  \AddToShipoutPictureFG*{%
    \AtPageUpperLeft{%
      \hspace*{0.93\textwidth}%
      \raisebox{-5.9cm}{\includegraphics[width=0.18\textwidth]{#1}}%
    }%
  }%
}

\title{Evaluating Time Awareness and Cross-modal Active Perception of Large Models via 4D Escape Room Task}

\setthunlparticlemeta{RESEARCH ARTICLE}
\thunlplogooff   

\author{
  \textbf{Yurui Dong\textsuperscript{3,*}}\quad
  \textbf{Ziyue Wang\textsuperscript{1,*}}\quad
  \textbf{Shuyun Lu\textsuperscript{4,*}}\quad
  \textbf{Dairu Liu\textsuperscript{5,*}}\quad
  \textbf{Xuechen Liu\textsuperscript{6,*}}\quad
  \textbf{Fuwen Luo\textsuperscript{1}}\quad
  \textbf{Peng Li\textsuperscript{2,$\dagger$}}\quad
  \textbf{Yang Liu\textsuperscript{1,2,$\dagger$}}\quad \\
  \thunlpauthoraffildivider}
  \thunlpaffiliation{
  \textsuperscript{1}Dept. of Comp. Sci. \& Tech., Institute for AI, Tsinghua University, Beijing, China\\
  \textsuperscript{2}Institute for AI Industry Research (AIR), Tsinghua University, Beijing, China\\
  \textsuperscript{3}Fudan University, Shanghai, China\\
  \textsuperscript{4}School of Computer \& Communication Engineering, University of Science and Technology Beijing, Beijing, China\\
  \textsuperscript{5}College of Software, Nankai University, Tianjin, China\\
  \textsuperscript{6}School of AI and Advanced Computing, Xi'an Jiaotong-Liverpool University, Jiangsu, China}

\date{March 2026}

\begin{document}

\addtitlesideimage{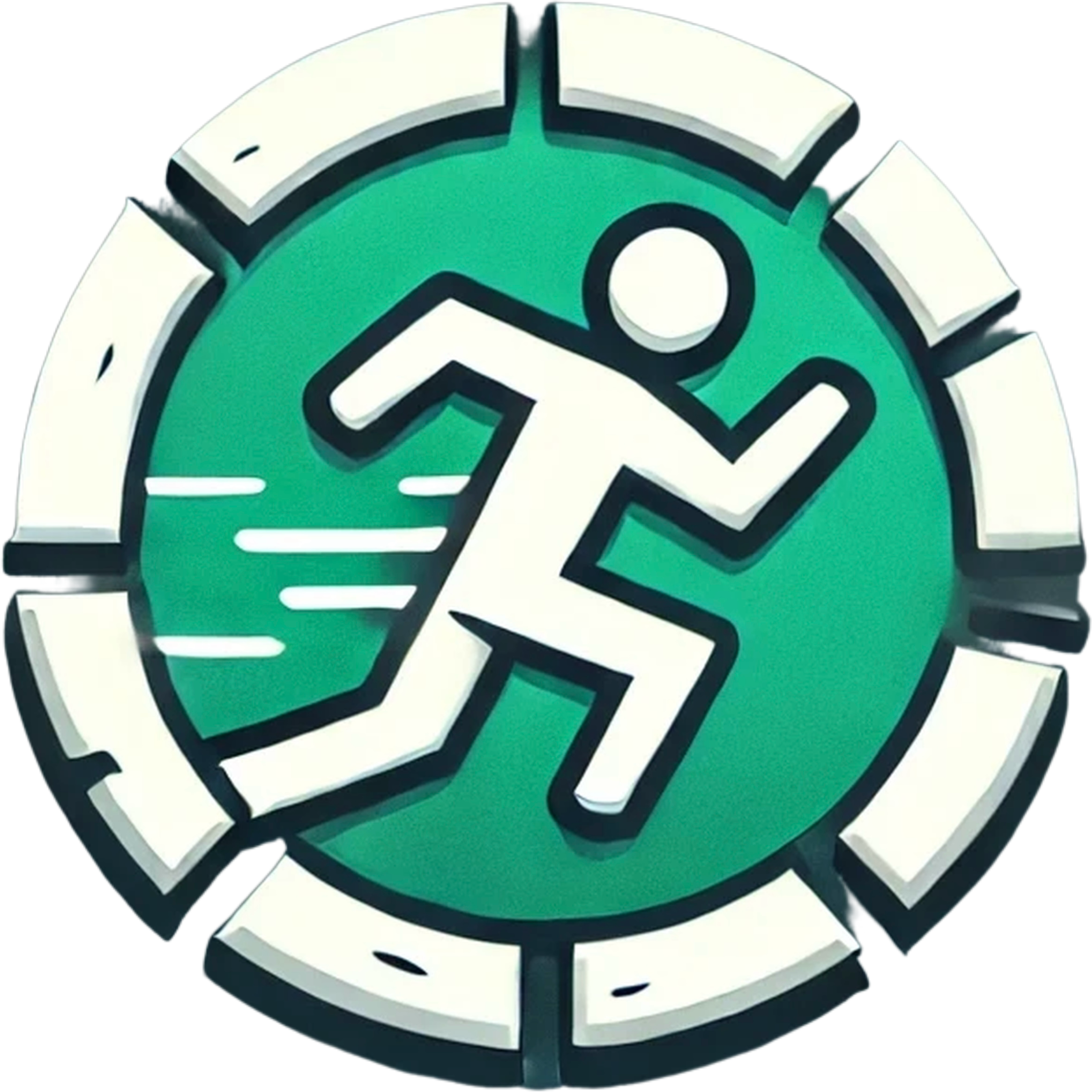} 
\makefrontmatter
{Multimodal Large Language Models (MLLMs) have recently made rapid progress toward unified Omni models that integrate vision, language, and audio. However, existing environments largely focus on 2D or 3D visual context and vision-language tasks, offering limited support for temporally dependent auditory signals and selective cross-modal integration, where different modalities may provide complementary or interfering information, which are essential capabilities for realistic multimodal reasoning. As a result, whether models can actively coordinate modalities and reason under time-varying, irreversible conditions remains underexplored. To this end, we introduce \textbf{EscapeCraft-4D}, a customizable 4D environment for assessing selective cross-modal perception and time awareness in Omni models. It incorporates trigger-based auditory sources, temporally transient evidence, and location-dependent cues, requiring agents to perform spatio-temporal reasoning and proactive multimodal integration under time constraints. Building on this environment, we curate a benchmark to evaluate corresponding abilities across powerful models. Evaluation results suggest that models struggle with modality bias, and reveal significant gaps in current model's ability to integrate multiple modalities under time constraints. Further in-depth analysis uncovers how multiple modalities interact and jointly influence model decisions in complex multimodal reasoning environments.\footnote{GitHub repo: \url{https://github.com/THUNLP-MT/EscapeCraft-4D}.}}

\begingroup
\renewcommand{\thefootnote}{\fnsymbol{footnote}}
\footnotetext[1]{Equal contribution.}
\footnotetext[2]{Corresponding author: pengli09@gmail.com,\\ liuyang2011@tsinghua.edu.cn.}
\endgroup

\input{sections/1.intro}

\input{sections/2.related}

\input{sections/3.env}
\input{sections/4.experiment}

\input{sections/5.analysis}

\input{sections/6.conclusion}

\bibliographystyle{thunlp_cvpr}
\bibliography{references}

\clearpage
\appendix
\input{sections/7.appendix}

\end{document}

%% file: sections/1.intro.tex
\section{Introduction}

\begin{figure*}[t]
    \centering
    \includegraphics[width=\textwidth]{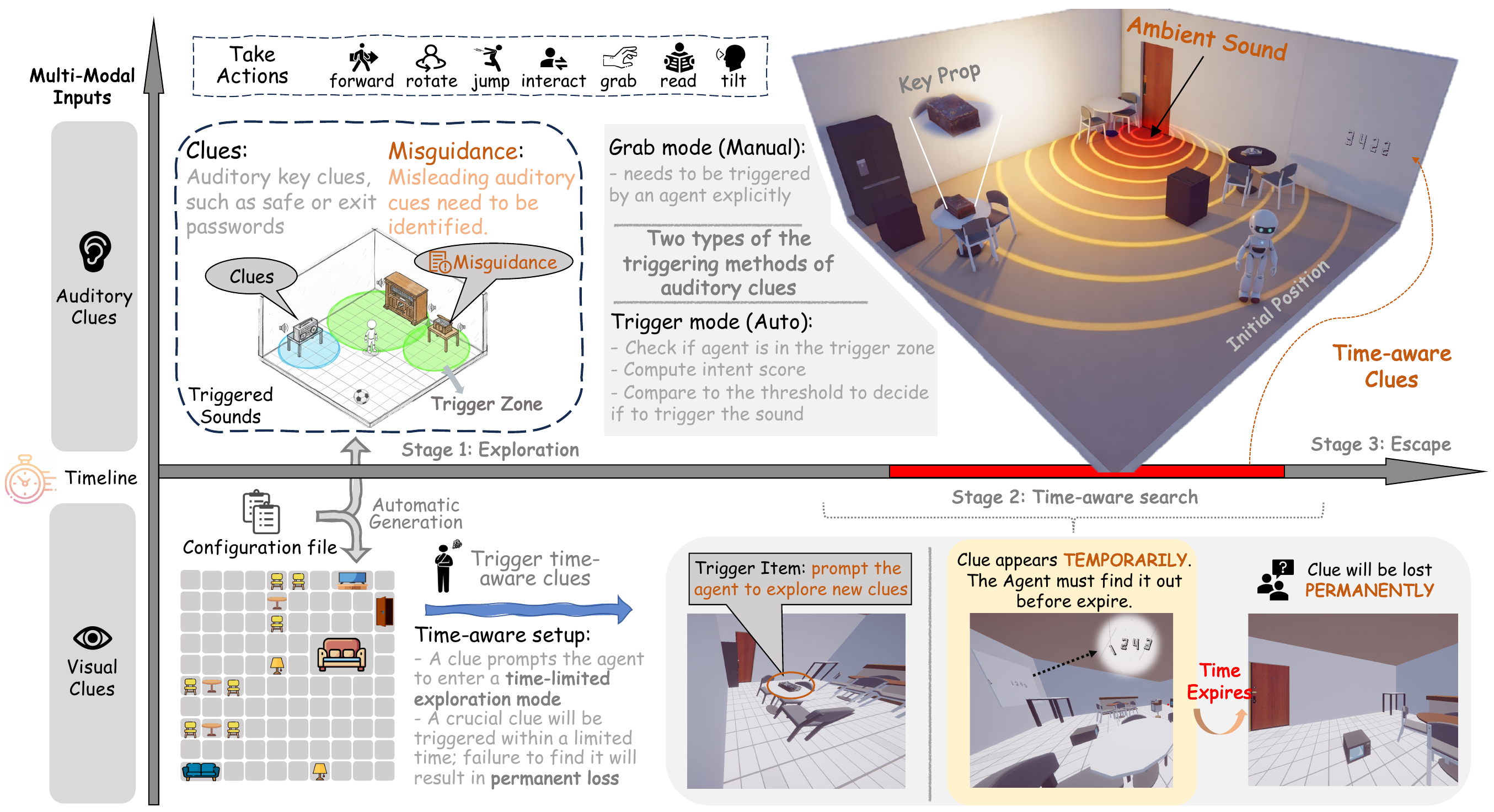}
    \caption{Our EscapeCraft-4D environment setup. It incorporates auditory, visual, and time-aware clues to evaluate multimodal reasoning. The system is designed to test the abilities of agents to coordinate cross-modal perception, such as leveraging spatially grounded auditory cues and visually presented clues or making time-aware decisions under strict temporal constraints. This figure illustrates the sequencially arranged tasks according to three stages: exploration, time-aware search, and finally escape, with dynamic auditory cues playing central roles in guiding the decision-making process. The trigger system dynamically introduces \textit{time-aware clues} that challenge the agent to act within limited timeframes, showcasing the integration of auditory and visual modalities.}
    \label{fig:main}
\end{figure*}

In recent years, Multimodal Large Language Models (MLLMs)~\cite{alayrac2022flamingo,liu2023Visual,o3o4systemcard,comanici2025Gemini,bai2025qwen3vltechnicalreport} have made rapid progress in effectively bridging heterogeneous perceptual modalities and enabling coherent reasoning across them, facilitating the application of MLLMs to a wide range of tasks, especially in Vision-Language (VL) including visual commonsense reasoning~\cite{zhou2024vicor, kil2024mllm}, spatial reasoning~\cite{yang2025thinking, zhan2025actial} and embodied manipulation and planning~\cite{driess2023palm, li2024manipllm}. Toward more realistic scenarios involving more than two modalities and time-aware interactions, prior work has commonly adopted agentic approaches that integrate isolated modules for different modalities~\cite{wu2023visual,shen2024hugginggpt,huang2023audiogpt}. More recent advances have shifted towards Omni models that process multiple modality streams within a unified architecture~\cite{xu2025qwen3omnitechnicalreport,wang2025MGMOmni,team2024chameleon, zhang2025streamomni}.

Despite this progress, the ability of current models to deeply integrate multiple modalities while preserving sensitivity to temporal signals remains underexplored. Existing evaluation paradigms focus primarily on static settings of 2D images or 3D volumetric scenes, often omitting the dimension of \textit{\textbf{time}}. This deficiency motivates the development of a 4-dimensional (4D) setup that incorporates dynamic and time-dependent factors to fully evaluate the capabilities to handle realistic tasks of Omni models. 
Furthermore, current multimodal benchmarks primarily emphasizes VL contexts, frequently relying on silent video clips and thereby overlooking the participation of auditory modality~\cite{li2023seed,li2024mvbench,fu2024videomme}. 
These benchmarks present two major shortcomings: 
First, they fall short on adequately evaluating \textit{\textbf{active perception across modalities}}~\cite{bajcsy1988active,chaplot2020learning,wilson2025popgs}, an essential capability that proactively determines \textit{when}, \textit{where}, and \textit{how} to selectively acquire multimodal information and suppress misleading information. 
Second, they overlook a critical aspect of real-life environments, \textit{\textbf{time awareness}} --- the ability to reason and act upon time-varying cues and irreversible information availability~\cite{wang-zhao-2024-tram, chen2025chronusomni}, beyond conventional temporal grounding of events to timestamps or intervals~\cite{gao2017tall,cai2024temporalbench}. Yet, dedicated evaluation and development of such time-aware reasoning, acting and planning capabilities remain limited in the current scope.

In this paper, we address the above limitations by introducing \textbf{EscapeCraft-4D}, a simulated environment that augments a 3D world with an explicit temporal dimension. This environment serves as a testbed for more realistic 4D tasks and advanced multimodal reasoning and acting. Specifically, we incorporate time awareness through time-varying visual cues and auditory signals, both of which naturally encode temporal information and require agents to reason under dynamic and irreversible conditions.

Moreover, EscapeCraft-4D is a highly customizable environment with more than two modalities, \textit{i.e.}, language, vision, and particularly audio. Notably, the inclusion of audio enables more complex design of cross-modal active perception where not all modalities are equally informative, requiring selective multimodal integration.
Our environment design establishes the 4D paradigm through two key mechanisms as visualised in Figure~\ref{fig:main}, specifically: 1) It is \emph{temporally dynamic}, requiring MLLMs to model temporal variations when aligning information across modalities; 2) It contains continuously active audio sources that provide location-dependent cues. Together, these properties require models to ground visual and auditory inputs within a unified spatio-temporal coordinate system and to selectively integrate multimodal information during decision making. Our contributions can be summarized as follows:
\begin{itemize}[left=0.4cm, itemsep=2pt, parsep=0pt]
    \item We introduce EscapeCraft-4D, a 4D environment that enables explicit evaluation of active cross-modal perception and time-aware reasoning in Omni models and agents.
    \item Our results reveal significant gaps in the current ability of Omni models to integrate multiple modalities under time constraints, including susceptibility to misleading auditory cues, limited time-aware interaction, and inconsistent cross-modal coordination across increasingly complex tasks.
    \item Detailed analysis demonstrates that auditory cues strongly influence exploration behavior, guiding agents toward relevant regions and enabling more effective multi-hop reasoning, while failures often correspond to missed temporal or spatial triggers, highlighting the critical role of cross-modal and time-aware coordination in complex multimodal tasks.
\end{itemize}

%% file: sections/2.related.tex
\section{Related Work}

\subsection{Multimodal Large Language Models}

Multimodal large language models (MLLMs) build on earlier vision-language models (VLMs) and audio-language models (ALMs). A representative early milestone is CLIP~\cite{radford2021Learning}, which learns aligned image-text representations via contrastive training in a shared embedding space. Subsequent work Flamingo~\cite{alayrac2022flamingo} augments a pretrained language model with visual conditioning modules, enabling in-context few-shot learning from interleaved image-text prompts. More recent open-source systems, BLIP~\cite{li2022BLIP}, LLaVA style visual instruction tuning~\cite{liu2023Visual}, MiniCPM-V~\cite{yao2024MiniCPMV}, and Qwen2.5-VL~\cite{bai2025Qwen25VLa}, further advance toward general purpose understanding and generation. Proprietary models such as GPT-4V~\cite{openai2024GPT4} and Gemini~\cite{geminiteam2024gemini15unlockingmultimodal} also demonstrate robust multimodal capabilities at scale. In parallel, ALMs condition language understanding and generation on audio signals, enabling models to \textit{hear} speech and general sounds~\cite{zhang2023SpeechLM, ghosh2024gamalargeaudiolanguagemodel, chu2024qwen2audiotechnicalreport, tang2024salmonngenerichearingabilities, kong2024audioflamingonovelaudio, zhang2023speechgptempoweringlargelanguage}. These efforts move toward unified language-centric frameworks that integrate heterogeneous perceptual modalities.

Recent research has further progressed toward an Omni paradigm that unifies vision, audio, and language with agentic pipeline~\cite{tao2026activeperceptionagentomnimodal} or within a single model (e.g., GPT-4o~\cite{openai2024gpt4o}, Qwen3-Omni~\cite{xu2025qwen3omnitechnicalreport}, and VITA-1.5~\cite{fu2025VITA15}). By jointly modeling these modalities, Omni models capture richer contextual cues, such as StreamOmni~\cite{zhang2025streamomni}, MGMOmni~\cite{wang2025MGMOmni}, LongCat-Flash-Omni~\cite{team2025LongCatFlashOmni}, and Gemini 2.5~\cite{comanici2025Gemini}. This trend represents a key step toward general purpose multimodal intelligence.

\subsection{Time-Aware Benchmarks and Perishable Evidence Collection}
In interactive and embodied reasoning tasks, information is frequently transient rather than persistent. In many scenarios, critical evidences and cues are triggered only after specific events and remains accessible for a limited time~\cite{zhou2024hazardchallengeembodieddecision}, e.g., a notification briefly appearing on a phone screen before disappearing. \cite{zhang2025etplanbenchembodiedtasklevelplanning} introduces causal dependency that ties decision quality to decision timing. Such constraint on sequential actions requires the agent to reason on both action selection and prioritization, where misaligned decisions on the sequence can lead to irreversible information loss and mistakes. Meanwhile, existing environments and benchmarks incorporate time mainly for action ordering~\cite{shridhar2020alfredbenchmarkinterpretinggrounded, srivastava2021behaviorbenchmarkeverydayhousehold, zhang2024plan}, where task-relevant information (e.g. actions required) remains stable rather than shaped by decision timing. The agent thus can defer exploration without lasting consequences, leaving time-aware reasoning and perishable evidence collection underexplored. In contrast, time-aware benchmarks evaluate whether agents recognize and respond to temporal changes in evidence availability, where delayed decisions can lead to irreversible information loss (e.g. a password that will disappear forever after a period of time).

%% file: sections/3.env.tex
\section{Environment and Task}
\subsection{Overview of EscapeCraft-4D}
\begin{figure*}[h]
    \centering
    \includegraphics[width=\textwidth]{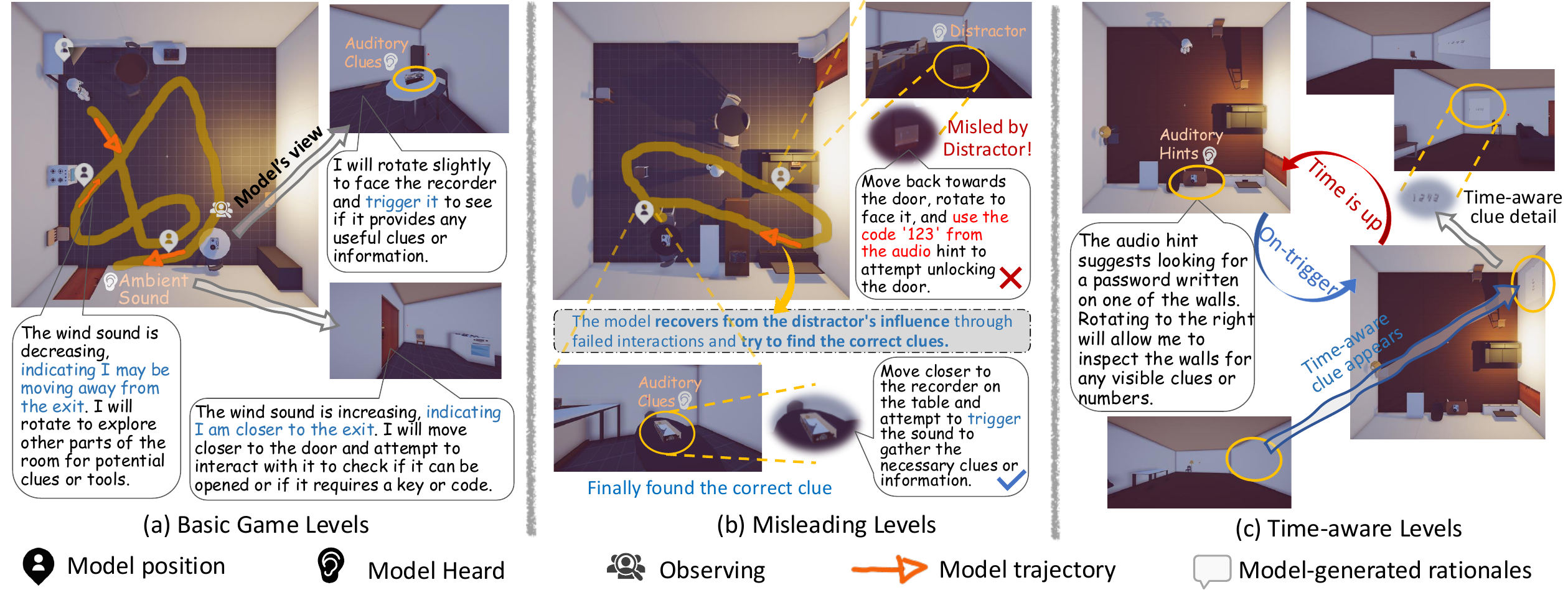}
    \caption{Examples of our designed levels with successful runs from GPT-4o. (a) Basic game design included both auditory and visual clues. (b) Misleading levels included distractors which would play misleading auditory clues. Models should distinguish which auditory clue was genuinely useful and which was the distractor. (C) Time-aware level design included clues only visible in limited time. Models should find these clues in the time limit or the clue would disappear.}
    \label{fig:case}
\end{figure*}

The room-escape task in this study is inspired by real-world escape room games and built on prior multimodal settings introduced in~\cite{wang2025multimodal, lim-etal-2025-visescape}. It is designed to evaluate the multimodal reasoning and planning capabilities of MLLMs. Following the 3D EscapeCraft environment~\cite{wang2025multimodal} that requires models to autonomously explore, perceive, and interact with the environment to achieve long-horizon goals, we introduce a fourth dimension, \textit{time}, by incorporating audio components as shown in Figure~\ref{fig:main}. Vision, language, and audio are three most commonly perceived and used modalities in human activities~\cite{stein1993merging,baltruvsaitis2018multimodal,liang2024foundations}. However, most current frameworks fail to address the time-awareness and lack mechanisms to integrate these modalities over time. Our integration of the audio signals and time-variant cues forces models to reason, act and planning within a specific duration, bridging the gap between static multimodal perception and dynamic, time-sensitive agency.

We especially include audio modality through two distinct categories, \emph{ambient audio} and \emph{trigger audio}. Specifically, following the task design protocol in~\cite{wang2025multimodal}, auditory cues are integrated directly into the prop chain so that each reasoning step may depend on both visual evidence and spatially grounded audio signals. To account for the temporal nature of these signals, we further expand the escape tasks to include time-variant cues, enabling the evaluation of time-awareness of models during acting and planning. This section details the integration of the audio modality and the design of corresponding tasks.

\subsection{Auditory Modality Integration}
\label{sec:auditory_clues}
\subsubsection{Ambient Audio} This audio signal is continuously emitted from a designated object (e.g., the sound of wind blowing outside the exit door) and spatially grounded to that object, as illustrated by the ``Ambient Sound'' example in the upper right part of Figure~\ref{fig:main}. Its perceived loudness is a deterministic function of the relative position from the agent to the source object, monotonically increasing as the agent approaches the source and attenuating with distance until inaudible beyond a predefined range. Although these sounds do not directly reveal symbolic content (e.g., passwords), they provide a gradient signal that can be exploited to (i) guide the direction of search, (ii) prioritize exploration targets, and (iii) coordinate visual attention with spatial movement. Thus, exploration shifts from purely visual scanning to selective cross-modal perception guided by audio.

\subsubsection{Trigger Audio} This represents the time-variant signal in our environment that enables the evaluation of time-awareness. It is bound to specific interactable objects (e.g., a recorder, as visualized in the ``Triggered Sounds'' example in the upper left part of Figure~\ref{fig:main}), and is activated only when the agent issues an explicit trigger request while remaining inactive for the rest of time. To make it a non-trivial prop-chain element rather than a information channel that can be triggered randomly and unexpectedly, activation is gated by physical conditions, the agent must satisfy proximity and orientation thresholds before interaction. Once triggered, the object plays a short speech message that may contain (a) password to or instruction for following steps or (b) deceptive content (e.g., plausible but irrelevant numbers). 


Compared to visually dominant setups~\cite{wang2025multimodal}, our audio- and time-augmented setting shifts the challenge from passive visual search to selective cross-modal perception and also has the ability to evaluate dynamic audio-visual fusion under deception. This design stresses (i) proximity-conditioned information acquisition, (ii) embodied verification of auditory claims, and 
(iii) strategic exploration policies that integrate continuous guide (ambient) and discrete prompt (triggered) auditory cues in the multi-hop reasoning.

\subsection{Time-aware Tasks and Levels}

We introduce time-aware levels where task-critical evidence (e.g., the password to unlock the exit) is explicitly time-varying or non-persistent rather than statically available throughout exploration. In these scenes, interacting with specific trigger objects (e.g., a recorder) initiates temporal events that cause evidence to appear, alter, or disappear within a limited time window. For instance, activating a recorder may reveal that a password will temporarily emerge on a wall for a limited time before vanishing permanently. This design enforces a strict coupling between perception, memory, and timing: \textit{agents must not only identify relevant interactions but also recognize the transient nature of the evidence and act before it expires.} By incorporating design elements such as passwords disappearing over short periods, the availability of evidence changes over time. These designs in the time-aware level go beyond temporal action ordering, and allow direct evaluation whether models can reason under irreversible, time-conditioned information loss.




\subsection{MM-Escape4D Benchmark}
\label{sec:benchmark}

\subsubsection{Basic game design} By incorporating the auditory clue designs described in Section~\ref{sec:auditory_clues}, we leverage the ``prop chain'' framework following MM-Escape~\cite{wang2025multimodal} to facilitate the batch generation of scenarios containing auditory modalities. 

Specifically, by treating auditory clues as an additional reasoning hop, we define three standardized difficulty levels for each room:
\begin{itemize}
    \item \textbf{Difficulty-1:} The single-hop reasoning path, requiring no visual props to escape. Models can exit by identifying and interacting directly with the door. 
    \item \textbf{Difficulty-2:} This level adds one additional reasoning hop compared to Difficulty-1. The model must additionally locate a specific sound-emitting object and trigger the audio to obtain the password, which is then used to unlock the door (example in Figure~\ref{fig:case}(a))
    \item \textbf{Difficulty-3:} A reasoning path with one additional hop beyond Difficulty-2 is required. It asks for both auditory clues and physical items (e.g., keys or passwords). This level integrates visual and auditory cues to evaluate the cross-modal synthesis capabilities of models.
\end{itemize}

\subsubsection{Misleading level design} Furthermore, to evaluate the robustness of models against interference, we particularly design ``misleading'' versions for Difficulty-2 and Difficulty-3, named \textbf{Difficulty-2-M} and \textbf{Difficulty-3-M}. While maintaining the same hop-wise structure, we introduce a parallel distractor sound source to the hop containing auditory information. This distractor plays irrelevant audio containing numerical data intended to confuse the judgement from model, as displayed by Figure~\ref{fig:case} (b). Misleading is measured by the ability of model to disregard this irrelevant information; if the model attempts to utilize the content of distractor, it is considered misled.

\subsubsection{Time-aware level design} Time-aware levels, denoted as \textbf{Difficulty-2-T}, extend Difficulty-2 by introducing perishable clues, which are time-variant and action-related. Specifically, the first hop in this level involves an object, such as recorder, that provides audio instructions guiding the agent to find the password within the subsequent constrained durations. This second hop is time constrained, requiring model to be aware of this limited duration. As illustrated in Figure~\ref{fig:main} (c), after the recorder is triggered, the password to the exit appears at a specified position only temporarily before permanently disappearing. To realistically simulate the passage of time within our environment, we assign specific time costs and corresponding speeds to all actions: linear velocity for movement, angular velocity for rotations (turning or looking up/down), and fixed time intervals for grabbing and triggering interactions. Detailed configurations are shown in Table~\ref{tab:time_cost}.

\begin{table}[h]
\caption{Time-cost Configuration and Audio Categories}
    \label{tab:time_cost}
    \centering
            \begin{tabular}{ll}
                \toprule
                \textbf{Category} & \textbf{Specification / Value} \\
                \midrule
                Forward Speed & 2.0 m/s \\
                Rotation Speed & 60$^\circ$/s \\
                Grab/Trigger & 0.5 s (fixed cost) \\
                \midrule
                \textbf{Audio Types} & \textbf{Examples} \\
                \midrule
                Ambient Sound & Wind sound from exit \\
                Auditory Clues & Passwords or prompts \\
                Misleading Audio & Distractors \\
                \bottomrule
            \end{tabular}
\end{table}

\subsubsection{Statistics of MM-Escape4D} For basic room settings, we generated 11 scenes for all the basic difficulties. For misleading levels, we generated 11 scenes for each of Difficulty-2-M and Difficulty-3-M. We also generated 11 scenes for Difficulty-2-T. There are totally 66 scenes for standard evaluation for auditory and visual abilities. Detailed statistics, including scenes and involved objects, are listed in Table~\ref{tab:statistics}. We also provide examples for each of our employed audio type in Table~\ref{tab:time_cost}.

\begin{table}[h]
    \caption{Statistical Summary of MM-Escape4D}
    \label{tab:statistics}
    \centering
            \begin{tabular}{lccc}
                \toprule
                \textbf{Level Type} & \textbf{Scenes} & \textbf{Total Objs} & \textbf{Avg/Scene} \\
                \midrule
                Difficulty-1 & 11 & 152 & 13.82 \\
                Difficulty-2 & 11 & 151 & 13.73 \\
                Difficulty-3 & 11 & 186 & 16.91 \\
                Difficulty-2-M & 11 & 188 & 17.09 \\
                Difficulty-3-M & 11 & 192 & 17.45 \\
                Difficulty-2-T & 11 & 163 & 14.82 \\
                \midrule
                \textbf{Total} & \textbf{66} & \textbf{1032} & \textbf{15.64} \\
                \bottomrule
            \end{tabular}
\end{table}

%% file: sections/4.experiment.tex
\section{Experiments}

\begin{table*}[t]
    \centering
    \caption{Results of basic game setting. Prop: Prop Gain; Steps: average steps used to complete the game; GSR: Grab Success Rate, the precision of grabbing; GR: Grab Ratio, the portion of grabbing actions regarding the total consumed steps. TSR: Trigger Success Rate. TR: Trigger Ratio. Note that Difficulty-1 and Difficulty requires no prop, and the prop gain is therefore omitted, and so as TSR and TR 
    The max allowed steps are 50, 65, 80 for Difficulty-1, -2, -3 respectively. The best score of each metrics is \textbf{bolded} and the second is \underline{underlined}. The GPT-4o Omni Agent refers to our agentic approach built on GPT-4o models, \texttt{gpt-4o} for vision and language, together with \texttt{gpt4o-audio-preview} for audio. We use Qwen3-Omni-30B-A3B version for Qwen3-Omni models in this table.}
    \label{tab:rst}
    \resizebox{\textwidth}{!}{
        \begin{tabular}{@{\hspace{0.1cm}}l|cccc|cccccc|ccccccc|c}
        \toprule 
         \multirow{2}{*}[-1.5ex]{Models} & \multicolumn{4}{c|}{Difficulty-1} & \multicolumn{6}{c|}{Difficulty-2} & \multicolumn{7}{c|}{Difficulty-3} & \multirow{2}{*}[-1.5ex]{{\begin{tabular}[c]{@{}c@{}}AVG\\ER (\%)$\uparrow$\end{tabular}}} \\ \cmidrule(lr){2-5} \cmidrule(lr){6-11}\cmidrule(lr){12-18} 
         & {\begin{tabular}[c]{@{}c@{}}ER\\(\%)$\uparrow$\end{tabular}} & Steps$\downarrow$ & {\begin{tabular}[c]{@{}c@{}}GSR \\ \small (\%)$\uparrow$\end{tabular}} & {\begin{tabular}[c]{@{}c@{}}GR\\ \small (\%)\end{tabular}} & 
         {\begin{tabular}[c]{@{}c@{}}ER\\(\%)$\uparrow$\end{tabular}} & Steps$\downarrow$ &  {\begin{tabular}[c]{@{}c@{}}GSR \\ \small (\%)$\uparrow$\end{tabular}} & {\begin{tabular}[c]{@{}c@{}}GR\\ \small (\%)\end{tabular}} & {\begin{tabular}[c]{@{}c@{}}TSR \\ \small (\%)$\uparrow$\end{tabular}} & {\begin{tabular}[c]{@{}c@{}}TR\\ \small (\%)\end{tabular}} & 
         {\begin{tabular}[c]{@{}c@{}}ER\\\small (\%)$\uparrow$\end{tabular}} & {\begin{tabular}[c]{@{}c@{}}Prop\\\small (\%)$\uparrow$\end{tabular}} & Steps$\downarrow$ &  {\begin{tabular}[c]{@{}c@{}}GSR\\ \small (\%)$\uparrow$\end{tabular}} & {\begin{tabular}[c]{@{}c@{}}GR\\ \small (\%)\end{tabular}} & {\begin{tabular}[c]{@{}c@{}}TSR \\ \small (\%)$\uparrow$\end{tabular}} & {\begin{tabular}[c]{@{}c@{}}TR\\ \small (\%)\end{tabular}} &  \\
         \midrule\noalign{\vskip -3pt}
        \multicolumn{19}{c}{\cellcolor[HTML]{EFEFEF} \textit{Agentic Approach}}    \\ 
        GPT-4o Omni-Agent & \cellcolor{green!20}\underline{90.91} & \cellcolor{green!20}\textbf{16.82} & \cellcolor{green!20}\underline{48.52} & \cellcolor{green!20}24.58 & \cellcolor{yellow!20}\textbf{81.82} & \cellcolor{yellow!20}\underline{38.82} & \cellcolor{yellow!20}\underline{12.75} & \cellcolor{yellow!20}33.66 & \cellcolor{yellow!20}55.30 & \cellcolor{yellow!20}9.15 & \cellcolor{orange!20}\textbf{45.45} & \cellcolor{orange!20}\textbf{45.45} & \cellcolor{orange!20}\underline{59.27} & \cellcolor{orange!20}\textbf{8.39} & \cellcolor{orange!20}34.73 & \cellcolor{orange!20}34.85 & \cellcolor{orange!20}24.09 & 72.73 \\
        \multicolumn{19}{c}{\cellcolor[HTML]{EFEFEF} \textit{Omni Models}}    \\ 
        
        \noalign{\vskip 1pt}

        Gemini3-Pro-Preview & \cellcolor{green!20}\textbf{100.00} & \cellcolor{green!20}\textbf{16.82} & \cellcolor{green!20}\textbf{59.25} & \cellcolor{green!20}21.95 & \cellcolor{yellow!20}\underline{72.73} & \cellcolor{yellow!20}\textbf{27.73} & \cellcolor{yellow!20}\textbf{37.12} & \cellcolor{yellow!20}22.67 & \cellcolor{yellow!20}\textbf{81.82} & \cellcolor{yellow!20}6.42 & \cellcolor{orange!20}\underline{27.27} & \cellcolor{orange!20}\underline{27.27} & \cellcolor{orange!20}\textbf{50.36} & \cellcolor{orange!20}\underline{7.82} & \cellcolor{orange!20}29.69 & \cellcolor{orange!20}\textbf{55.11} & \cellcolor{orange!20}4.52 & 66.67 \\
        
         Qwen3-Omni-Thinking & \cellcolor{green!20}63.64 & \cellcolor{green!20}\underline{24.00} & \cellcolor{green!20}24.80 & \cellcolor{green!20}54.97 & \cellcolor{yellow!20}27.27 & \cellcolor{yellow!20}56.73 & \cellcolor{yellow!20}5.50 & \cellcolor{yellow!20}29.97 & \cellcolor{yellow!20}39.86 & \cellcolor{yellow!20}22.24 & \cellcolor{orange!20}0.00 & \cellcolor{orange!20}0.00 & \cellcolor{orange!20}81.00 & \cellcolor{orange!20}0.00 & \cellcolor{orange!20}27.27 & \cellcolor{orange!20}\underline{42.29} & \cellcolor{orange!20}19.53 & 30.30 \\
         
         Qwen3-Omni-Instruct & \cellcolor{green!20}18.18 & \cellcolor{green!20}47.55 & \cellcolor{green!20}6.82 & \cellcolor{green!20}9.24 & \cellcolor{yellow!20}27.27 & \cellcolor{yellow!20}60.27 & \cellcolor{yellow!20}12.21 & \cellcolor{yellow!20}7.99 & \cellcolor{yellow!20}\underline{56.06} & \cellcolor{yellow!20}2.67 & \cellcolor{orange!20}0.00 & \cellcolor{orange!20}0.00 & \cellcolor{orange!20}81.00 & \cellcolor{orange!20}0.00 & \cellcolor{orange!20}9.09 & \cellcolor{orange!20}29.52 & \cellcolor{orange!20}6.06 & 15.15 \\

         MGM-Omni-32B & \cellcolor{green!20}27.27 & \cellcolor{green!20}39.55 & \cellcolor{green!20}10.91 & \cellcolor{green!20}17.14 & \cellcolor{yellow!20}0.00 & \cellcolor{yellow!20}66.00 & \cellcolor{yellow!20}0.00 & \cellcolor{yellow!20}17.08 & \cellcolor{yellow!20}14.55 & \cellcolor{yellow!20}1.10 & \cellcolor{orange!20}0.00 & \cellcolor{orange!20}0.00 & \cellcolor{orange!20}81.00 & \cellcolor{orange!20}0.00 & \cellcolor{orange!20}17.51 & \cellcolor{orange!20}18.69 & \cellcolor{orange!20}2.81 & 9.09 \\
         \bottomrule 
         
    \end{tabular}
     } 
\end{table*}

\begin{table*}[t]
    \centering
    \caption{Comprehensive Experimental Results on MM-Escape4D. (I) Performance on \textbf{Time-aware Levels} measured by TCSS. (II) Robustness against \textit{Misleading Auditory} cues: \textbf{MAT} refers to ratio of scenes where \textbf{m}isleading \textbf{a}udio was \textbf{t}riggered; \textbf{AMR} represents the ratio of scenes where the agent was deceived given that the misleading cue was triggered, within 3 following steps. \textbf{Dn-M-ER} and \textbf{Dn-M-Step} denote the Escape Rate and average steps for Difficulty-n Misleading levels. The best and second-best results are highlighted in \textbf{bold} and \underline{underline}, respectively.}
    \label{tab:combined_results}
    \resizebox{\textwidth}{!}{
        \begin{tabular}{l c cccccccc} 
            \toprule
            \multirow{2}{*}{\textbf{Models}} & \textbf{Time-Aware} & \multicolumn{8}{c}{\textbf{Misleading Levels}} \\
            \cmidrule(lr){2-2} \cmidrule(lr){3-10}
             & \textbf{TCSS$\uparrow$} & \textbf{MAT$\downarrow$} & \textbf{AMR$\downarrow$} & \textbf{D2-M-ER$\uparrow$} & \textbf{D3-M-ER$\uparrow$} & \textbf{D2-M-Steps$\downarrow$} & \textbf{D3-M-Steps$\downarrow$} & & \\ 
            \midrule
            \textit{Agentic Approach} & & & & & & & & & \\
            GPT-4o Omni-Agent & \textbf{67.95\%} & 63.64\% & 100.00\% & \textbf{72.73\%} & \textbf{63.64\%} & \underline{40.55} & \underline{60.09} & & \\
            \midrule
            \textit{Omni Models} & & & & & & & & & \\
            Qwen3-Omni-Thinking & \underline{55.87\%} & 59.09\% & 23.08\% & \phantom{0}0.00\% & \phantom{0}0.00\% & 66.00 & 81.00 & & \\
            Gemini-3-Pro-Preview & 47.04\% & \underline{40.91\%} & \underline{22.22\%} & \underline{63.64\%} & \textbf{63.64\%} & \textbf{28.82} & \textbf{53.82} & & \\
            MGM-Omni-32B & \phantom{0}0.00\% & \textbf{13.64\%} & \textbf{\phantom{0}0.00\%} & \phantom{0}0.00\% & \phantom{0}0.00\% & 66.00 & 81.00 & & \\
            Qwen3-Omni-Instruct & \phantom{0}0.00\% & \underline{40.91\%} & \textbf{\phantom{0}0.00\%} & \phantom{0}0.00\% & \phantom{0}0.00\% & 66.00 & 81.00 & & \\
            \bottomrule
        \end{tabular}
    }
\end{table*}

\subsection{Evaluation Metrics}

For the assessment of model capabilities and behaviors, we follow~\cite{wang2025multimodal} and adopt metrics including, Escape Rate (ER), Prop Gain, Grab Success Rate (GSR), Grab Ratio ($\text{R}_{grab}$), and Intent-Outcome Consistency ($C_{IO}$). Please refer to Appendix~\ref{app:metric} for detailed calculation and explanation. 

In addition to these well-defined metrics, we design four additional metrics dedicated to auditory signals, cross-modal active perception, and time-awareness in this paper, to address the targets of our new 4D environment. Specifically, we introduce (i) \textbf{A}udio \textbf{M}isguidance \textbf{R}ate (AMR) to quantify how often models are distracted by irrelevant modality, (ii) \textbf{T}ime-\textbf{C}onstrained \textbf{S}earch \textbf{S}core (TCSS) to measure how early a model finds the target relative to the time limit, with zero score assigned if the target is not found within the time budget. (iii) \textbf{T}rigger \textbf{r}atio (TR) and (iv) \textbf{T}rigger \textbf{s}uccess \textbf{r}atio (TSR). Denoting the total steps as $S$, the number of scenes in which the model is misled by irrelevant audio cues as $N_{\text{misguided}}$, amount of succeeded trigger action as $N_{tigger}^{TP}$, trigger success rate as $\text{TSR}$, trigger ratio as $\text{R}_{trigger}$, and $N_{\text{total}}$ the number of scenes where misleading cue was triggered. Let $t_{\text{found}}$ denote the consumed time duration when the target is found, and $T_{lim}$ the time limit for each episode. We have,

\begin{equation}\label{eq:1}
        \text{TSR} = \frac{N_{trigger}^{TP}}{\sum \text{Triggering actions}},
\end{equation}
\begin{equation}\label{eq:2}
\text{R}_{trigger} = \frac{\sum \text{Triggering actions}}{S},
\end{equation}
\begin{equation}\label{eq:3}
    \text{AMR} = \frac{N_{\text{misguided}}}{N_{\text{total}}},
\end{equation}
\begin{equation}\label{eq:4}
    \text{TCSS} =
    \begin{cases}
        1 - \dfrac{t_{\text{found}}}{T_{lim}}, & \text{if found within } T_{\lim}, \\
        0, & \text{otherwise}.
    \end{cases}
\end{equation}

\subsection{Evaluation Setup}
We evaluated a range of open-source and proprietary models of varying scales, utilizing the tiered methodology described in Section~\ref{sec:benchmark} to assess their multimodal reasoning capabilities. Open-source models include MGM-Omni-32B~\cite{wang2025MGMOmni}, Qwen3-0mni-30B-A3B-Thinking and Qwen3-Omni-30B-A3B-Instruct~\cite{xu2025qwen3omnitechnicalreport}, proprietary models include GPT-4o~\cite{openai2024gpt4o} and Gemini-3-Pro~\cite{geminiroboticsteam2025geminiroboticsbringingai}. For GPT-4o, we implement an agentic approach to enable Omni ability, as the officially provided APIs do not support simultanuously inputting all three modalities, text, image, and audio. Specifically, we employ \texttt{gpt-4o} as the VL backbone, and equip it with \texttt{gpt4o-audio-preview} to process any auditory information when required.

Detailed evaluation templates are provided in Appendix~\ref{app:prompt}. Following the experimental design in~\cite{wang2025multimodal}, we set the maximum number of interaction steps to 50, 65, and 80 for Difficulty-1, Difficulty-2, and Difficulty-3, respectively, including their misleading and time-aware variants. This configuration ensures that models are granted sufficient opportunities to explore and interact with the environment, while preventing excessively long trajectories that could obscure performance differences across difficulty levels.

\subsection{Experimental Results}
\subsubsection{Results of Basic Game Settings}

\begin{figure*}[t]
    \centering
    \includegraphics[width=\textwidth]{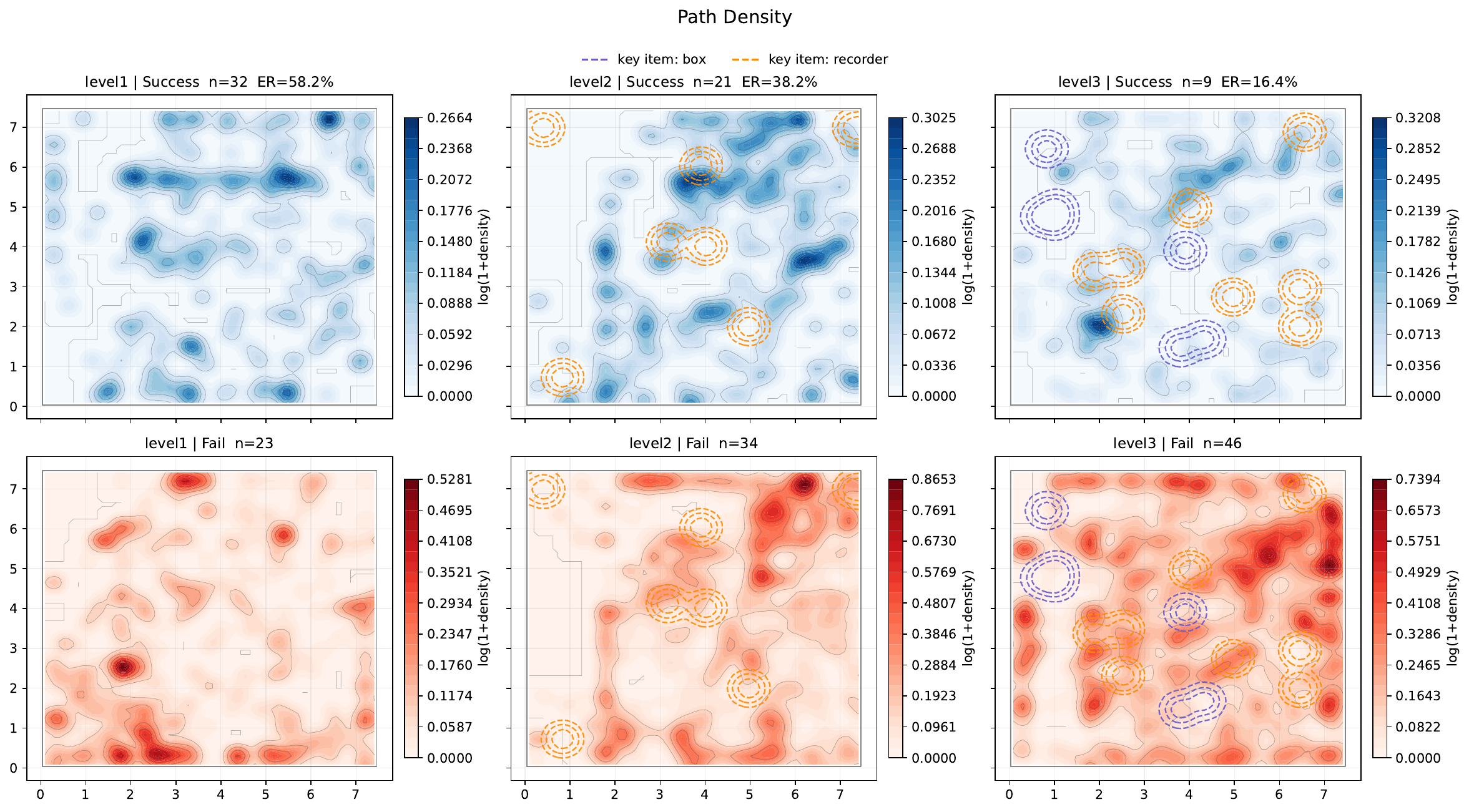}
    \caption{\textbf{Path density analysis for different levels in our MM-Escape4D benchmark.} Exits of different scenes are aligned as mentioned in Section~\ref{sec:ana_path}
The heatmaps represent the log-normalized density ($\log(1+\text{density})$) of agent positions, with blue gradients for successful runs and red for failed ones. 
The dashed contours represent the spatial distribution of key items (e.g., recorders for auditory clues and boxes). }
    \label{fig:density}
\end{figure*}

The evaluation results in Table~\ref{tab:rst} reveal a substantial and widening performance gap as task difficulty increases. GPT-4o Omni-agent achieves an average ER of 72.73\%, followed by Gemini-3-Pro (66.67\%), both more than doubled of the strongest open-source model, Qwen3-Omni-Thinking (30.30\%). The gap becomes more distinct at Difficulty-3, where all open-source candidates fail to escape, while proprietary models remain capable of completing games.
This discrepancy suggests that MM-Escape4D measures capabilities not well captured by conventional multimodal benchmarks. Unlike those tasks, our task requires models to actively explore 3D space, trigger object-conditioned auditory cues, and reason over transient and irreversible signals (e.g., time-limited clues that permanently disappear), demanding tight coupling between perception, reasoning, and action in a time-sensitive interactive setting.

Moreover, strong performance on conventional and common multimodal benchmarks does not necessarily translate to embodied and real-world multimodal decision making. Although recent open-source Omni models report competitive results on widely used benchmarks, such as MMMU~\cite{yue2024mmmu}, MathVista~\cite{lu2024mathvista}, and AIME25\footnote{\scriptsize{American Invitational Mathematics Examination (AIME) 2025, \url{https://maa.org}.}}, they degrade substantially in our EscapeCraft-4D. Qwen3-Omni-Thinking drops to 27.27\% ER on Difficulty-2 and 0\% on Difficulty-3, implying persistent limitations in spatial reasoning and cross-modal coordination, particularly in time-sensitive and modality-intensive scenarios

Finally, the overall results suggest structural advantages for unified multimodal modeling. While current Omni models still struggle with effective modality selection during interactive decision making, they demonstrate better modality coordication than the agentic approach, GPT-4o Omni-agent, where audio is first converted into text before reasoning, reducing the task to VL setting. In practice, our evaluated Omni models, including the open-source Qwen3-Omni, achieve higher TSR than GPT-4o agent, indicating that integrating inputs of multiple modalities into unified form can help supports complex cross-modal reasoning.


\subsubsection{Results of Misleading Modality and Time-aware Settings}

\begin{figure*}[tb]
    \centering
    \includegraphics[width=\textwidth]{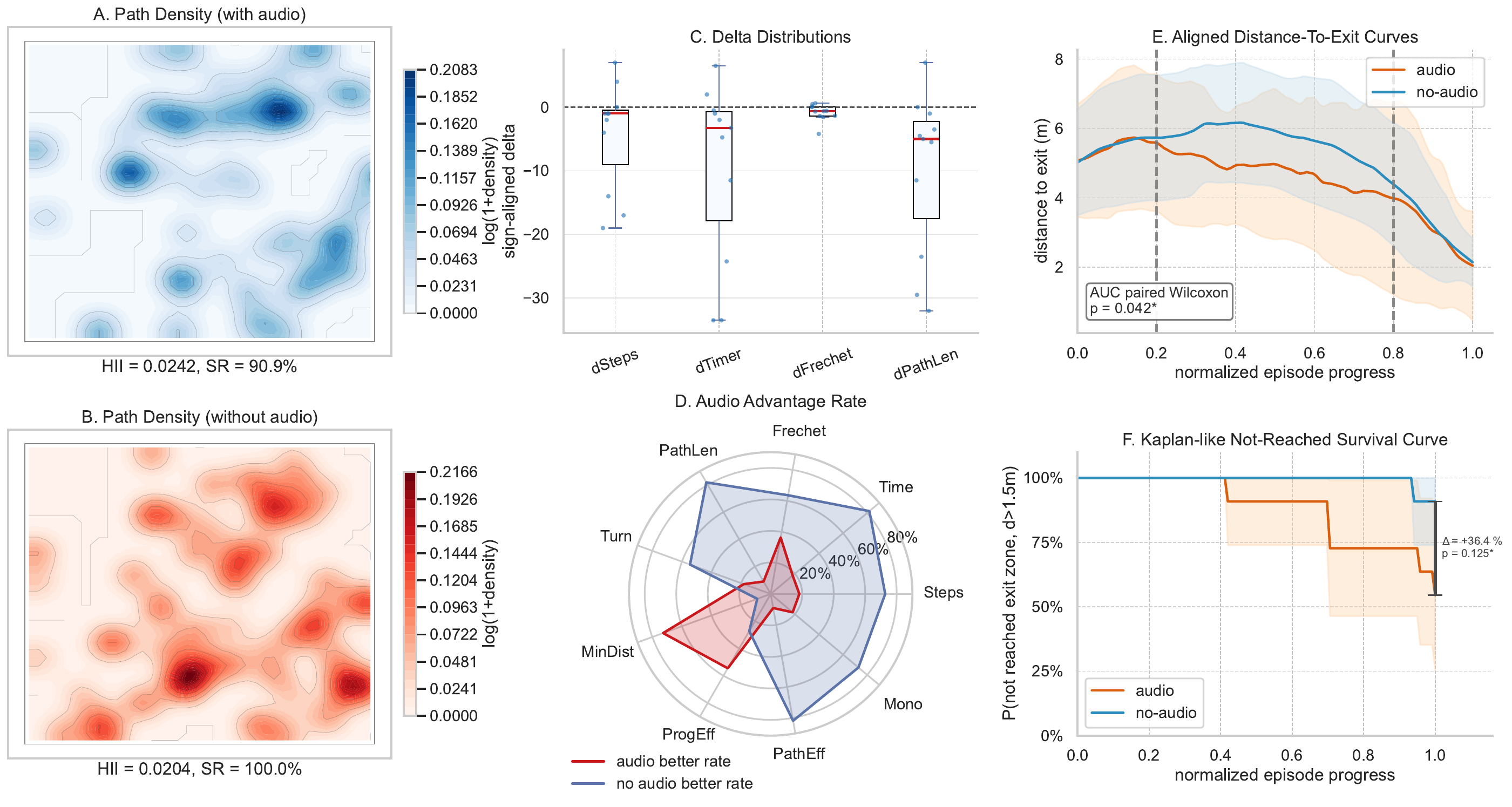}
    \caption{\textbf{Analysis of Ambient Sound via Audio and No-Audio settings.} 
(A-B) Path density heatmaps showing trajectory concentrations, using the same spatial aligning method as in Figure~\ref{fig:density}. 
(C) Delta distributions for sign-aligned metrics (\textit{Steps, Timer, Frechet, PathLen}); the red median lines below zero indicate the superiority of the audio condition. 
(D) The winning rate of audio guidance across nine navigation metrics. Metrics where lower values are better: \textit{Steps}, \textit{Time}, \textit{PathLen}, \textit{Turn}, and \textit{Frechet}. Efficiency metrics where higher values are preferred: \textit{ProgEff}, \textit{PathEff}, and \textit{Mono}.
(E) Mean distance-to-exit over normalized episode progress; the shaded area represents $\pm 1$ standard deviation. 
(F) Kaplan-like survival plot showing the percentage of models that have not yet reached the 1.5m exit threshold. 
Statistical significance for AUC and endpoint success is determined via paired Wilcoxon tests ($*p < 0.05$ indicates a statistically significant difference).}
    \label{fig:level1}
\end{figure*}

In misleading modality settings, robustness is governed by both exposure and susceptibility: MAT (misleading cue triggered) and AMR (deceived given trigger). GPT-4o exhibits high exposure and susceptibility (MAT $63.64\%$, AMR $100.00\%$), whereas Qwen3-Omni-Thinking has similarly high exposure (MAT $59.09\%$) but lower susceptibility (AMR $23.08\%$), though its misleading-level ER remains $0.00\%$. Gemini achieves the best balance 
among the reported metrics. Other open-source models (MGM-Omni-32B, Qwen3-Omni-Instruct) remain near-zero on misleading completion (both D2-M-ER and D3-M-ER at $0.00\%$), indicating interaction failures before cross-modal disambiguation can occur.

For the time-aware setting, TCSS directly measures how quickly a model finds the correct transient clue within the 20s window. TCSS differs substantially across models: GPT-4o achieves the highest score ($67.95\%$), followed by Qwen3-Omni-Thinking ($55.87\%$) and Gemini-3-Pro-Preview ($47.04\%$), while MGM-Omni-32B and Qwen3-Omni-Instruct remain at $0.00\%$. Across all runs, higher TCSS is significantly associated with final success (Spearman $r=0.433$, $p=9.56\times10^{-4}$), and successful runs show significantly higher TCSS than failed runs (permutation test, $p=0.0016$). These results reveal the gap that strong performance on static multimodal benchmarks does not directly transfer to embodied, time-variant cross-modal decision making. Detailed results are shown in Table~\ref{tab:combined_results}.

Overall, we summarise the following takeaways:
\begin{itemize}
    \item Active perception under irreversible constraints exposes fundamental weaknesses. When agents must explore, trigger object-conditioned auditory cues, and reason over transient information, current Omni models exhibit pronounced degradation.
	\item 3D spatial reasoning and cross-modal temporal integration remain underdeveloped. Coordinating vision, language, and audio in dynamic settings is significantly more challenging than solving static or time-invariant multimodal reasoning tasks.
	\item Unified modality modeling improves coordination efficiency. Modality fusion demonstrates advantages in cross-modal interaction over modular, transcription-based agentic pipelines, indicating that structural integration of input modalities is beneficial for complex and dynamic decision making.
\end{itemize}

%% file: sections/5.analysis.tex
\section{Analysis}

\begin{figure*}[tb]
    \centering
    \includegraphics[width=\textwidth]{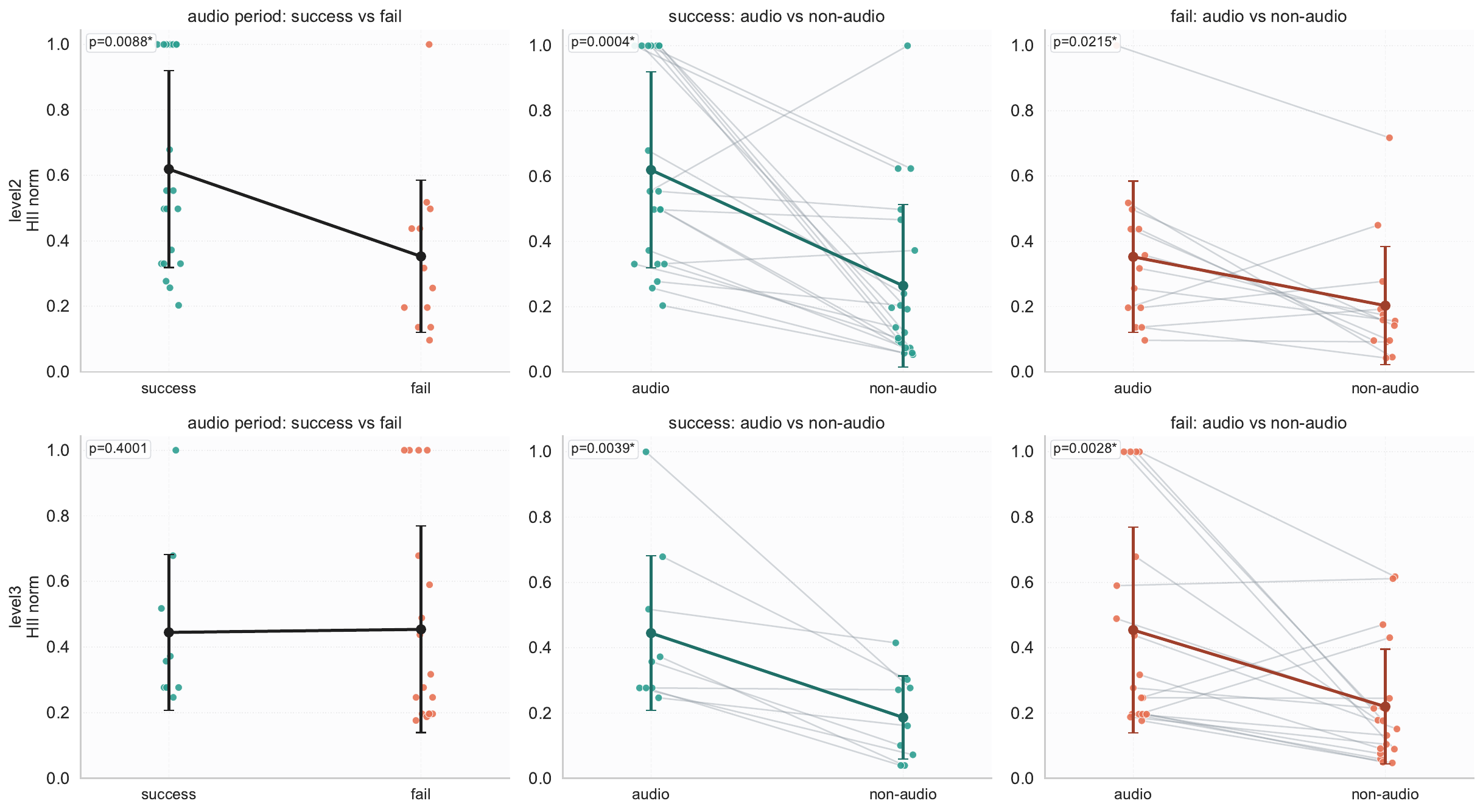}
    \caption{\textbf{Paired HII comparison under auditory conditioning for Difficulty-2 and Difficulty-3.}
Each row corresponds to one difficulty level.
Left column: unpaired success vs fail comparison in audio-active segments.
Middle and right columns: paired audio vs non-audio comparison within success and fail runs.
Each dot is one run-level \(\mathrm{HII}_{\mathrm{norm}}\); lines connect paired points from the same run; error bars show mean \(\pm\) standard deviation. The figure shows consistent concentration increase during audio-active periods, while success vs fail separation in audio periods is significant in Difficulty-2 and not significant in Difficulty-3.
}
    \label{fig:hii}
\end{figure*}

\begin{table*}[htb]
  \centering
  \caption{\textbf{P-values for the hypothesis tests used}.}
  \label{tab:hii_results2}
  \begin{tabular}{l cccc cc c}
      \toprule
    Test & \multicolumn{4}{c}{\textbf{HII}} & \multicolumn{2}{c}{\textbf{JS Distance}} & \multicolumn{1}{c}{\textbf{Visual Effects}} \\
    \cmidrule(r){2-5} \cmidrule(lr){6-7} \cmidrule(l){8-8}
    \textbf{Difficulty} & $P_{A \text{ vs } NA}^s$ & $P_{A \text{ vs } NA}^f$ & $P_{aud}^{s/f}$ & $P_{non-aud}^{s/f}$ & $P^{\mathrm{sf}}_{\mathrm{L_1},\,\mathrm{A\ vs\ NA},\,S}$ & $P^{\mathrm{sf}}_{\mathrm{JSD},\,\mathrm{A\ vs\ NA},\,F}$ & $P^{\mathrm{Mantel}}_{\mathrm{box\leftrightarrow path}}$  \\
    \midrule
    Difficulty-1 & -- & -- & -- & 0.3612 & -- & -- & -- \\
    Difficulty-2 & \textbf{0.0004*} & \textbf{0.0215*} & \textbf{0.0088*} & 0.2388 & \textbf{0.00033*} & \textbf{0.00033*} & -- \\
    Difficulty-3 & \textbf{0.0039*} & \textbf{0.0028*} & 0.4001 & 0.3568 & \textbf{0.00033*} & \textbf{0.00033*} & \textbf{0.0047*}\textbf{\phantom{*}} \\
    \bottomrule
    \addlinespace[1ex]
    \multicolumn{8}{l}{\footnotesize * Note: $P < 0.05$ indicates a statistically significant difference.}
  \end{tabular}
  \vspace{-5pt}
\end{table*}

\subsection{Analysis of Exploration Behavior and Path}~\label{sec:ana_path}
To investigate the underlying exploration behavior, navigation strategies, and the impact of cross-modal cues on decision-making, we visualize the movement patterns using exit-aligned path density heatmaps in Figure~\ref{fig:density}. 
We rotate each trajectory by $90^\circ$ increments so that the target exit is consistently aligned at the top edge for easier comparison, and then aggregate the spatial distributions of successful and failed runs across different level families.

In Difficulty-1, successful escape exhibit a direct and concentrated path towards the exit, whereas failed runs show sparse exploration, suggesting a lack of basic spatial grounding. 
As complexity increases in Difficulty-2 and -3, the density maps reveal a strategic shift, where models must deviate from the exit-ward path to interact with key items (e.g., recorders or boxes), as indicated by the dashed contour overlays.
Note that successful runs in these levels show higher density around these key items compared to failed runs, indicating more effective active perception and target-oriented exploration. The lower ER in more complex levels correlate with more fragmented and exploratory path distributions.

A significant observation in failed runs for Difficulty-2 and -3 is that these runs are often trapped near the center or far-flung corners, where models often wander aimlessly or fail to transition from auditory-based exploration to final escape. 
The discrepancy between success and failure is most pronounced in Difficulty-3, where successful runs demonstrate a clear multi-hop traversal pattern, e.g., first localizing the auditory clue (recorder) and then proceeding to the exit, while failed ones often miss the crucial temporal or spatial trigger zones of these objects.


\subsection{Ablation Study of Ambient Sound}

To analyze the specific impact of ambient sound on active perception, we conduct an ablation study using Difficulty-1, which is uniquely suited for the analysis as it requires no complex prop chains or specific clues (e.g., passwords) to escape, relying solely on the ability to navigate toward the exit. By comparing the standard Difficulty-1 setting with a spatialized ambient sound source at the exit with a silent version, the effect of auditory cues on spatial reasoning and decision-making under uncertainty can be isolated.

We conduct analysis across nine metrics, including burden-related measures such as \textit{Steps}, \textit{Time}, \textit{PathLen} (the total actual path length), \textit{Turn} (turning back toward the opposite direction), and \textit{Frechet}, as well as efficiency indicators including \textit{ProgEff} (Progress Efficiency), \textit{PathEff} (Path Efficiency), \textit{Mono} (Monotonicity Ratio of trajectory), and \textit{MinDist} (the closest proximity achieved to the exit). Together, these metrics characterize exploration behavior, path optimality, and progress toward the goal.  Detailed explanations on these measurements can be found in Appendix~\ref{app:foundations}.

\paragraph{Results and Analysis}
Figure~\ref{fig:level1} compares the results between the Audio and No-Audio conditions in Difficulty-1. It shows that spatialized ambient sound significantly improves active perception and navigational efficiency. As illustrated in the Delta Distributions (Figure~\ref{fig:level1} C), auditory cues leads to a substantial reduction in \textit{Steps}, \textit{Time}, \textit{PathLen}. Path quality is further evaluated using the \textit{Frechet}. The Non-Audio group exhibits consistently higher Frechet values compared to the Audio group, reflecting significantly more wandering and path distortion. Correspondingly, the median deltas ($\Delta = \text{Audio} - \text{No-Audio}$) for all these burden-related metrics fall well below the zero baseline, confirming that auditory feedback helps bypass redundant search behaviors and act with greater decisiveness and efficiency.

The performance gap is further highlighted by the Audio Advantage Radar Chart (Figure~\ref{fig:level1} D), which shows the win rate of audio-guided settings across nine metrics. In this chart, the audio group presents lower burden in burden-related metrics \textit{Steps}, \textit{Time}, \textit{PathLen}, \textit{Turn}, and \textit{Frechet}, and achieves higher efficiency in \textit{ProgEff}, \textit{PathEff}, and \textit{Mono}, suggesting a strong audio benefit. The high better-rate in \textit{MinDist} confirms that audio-guided agents not only move efficiently but also approach the exit more successfully. Overall, audio improves performance in over 60\% of trials across nearly all dimensions, demonstrating robust multi-faceted advantages for navigation.

Temporal behavior analysis using Aligned Distance-to-Exit curves (Figure~\ref{fig:level1} E) shows the Audio group consistently closer to the exit between the central portion of the episode (0.2–0.8), with lower variance than the No-Audio group. A Paired Wilcoxon Signed-Rank Test on the AUC confirms the statistical significance ($P^{\mathrm{AUC}}_{\mathrm{A\ vs\ NA}}=0.042^*<0.05$), indicating that models actively leverage auditory cues rather than random variation. Similar trends appear in Kaplan-like Not-Reached survival curves in Figure~\ref{fig:level1} F, where the Audio curve decays faster, implying earlier entry into the exit zone. Another Paired Wilcoxon Test on terminal non-arrival rates ($P^{\mathrm{AUC}}_{\mathrm{A\ vs\ NA}}=0.0125^*<0.05$) further confirms that our ambient sound accelerates the transition from exploration to goal-oriented navigation with significance. 

To test the comprehensive behavior through the whole path, a point-label premutation test was conducted on the path density. By deviding the room in to a $M \times M$ grid, we get the path density distribution $\mathcal{P}_{aud}$ and $\mathcal{P}_{non-aud}$. Since the levels of audio and non-audio are exactly the same, we can use both $L_1$ and $JS$ distance to calculate the gap between the path density of each level. The point-label premutation test reveals significant difference between the path density ($P^{\mathrm{perm}}_{\mathrm{L1}}=0.00019^*<0.05, P^{\mathrm{perm}}_{\mathrm{JSD}}=0.00019^*<0.05$), which confirms that the ambient sounds had a significant impact on the choice of agent path.

\subsection{Influence of Auditory Clues}
\newcolumntype{Y}{>{\raggedright\arraybackslash}X}
\begin{table*}[htb]
\centering
\caption{Tests used for analysis.}
\label{tab:testwise_mapping_ch5_style}
\begin{tabularx}{\textwidth}{>{\raggedright\arraybackslash}p{0.23\linewidth} 
                            >{\raggedright\arraybackslash}p{0.25\linewidth} 
                            Y}  
\toprule
Test Symbol & Test Type & What Is Being Tested \\
\midrule

$P^{\mathrm{AUC}}_{\mathrm{A\ vs\ NA}}$ &
Paired Wilcoxon signed-rank Appendix~\ref{sec:wilcoxon} &
Whether audio changes the mid-trajectory distance-to-exit behavior between Difficulty-1 audio and non-audio settings (AUC on normalized progress window). \\

$P^{\mathrm{end}}_{\mathrm{A\ vs\ NA}}$ &
Paired Wilcoxon signed-rank Appendix~\ref{sec:wilcoxon} &
Whether audio changes endpoint not-reached status between Difficulty-1 audio and non-audio settings (Kaplan-like endpoint comparison). \\

$P^{\mathrm{perm}}_{\mathrm{L1}}$, $P^{\mathrm{perm}}_{\mathrm{JSD}}$ &
Point-label permutation test Appendix~\ref{sec:perm} &
Whether the density between difficulty-1 audio/non-audio settings changes significantly using $L_1$ and $JS$ distance. \\

$P^{\mathrm{A\ vs\ NA}}_{\mathrm{S/F}}$ &
Paired Wilcoxon signed-rank Appendix~\ref{sec:wilcoxon} &
Within successful/failed runs, whether HII behaviors change among audio/non-audio segments. \\

$P^{s/f}_{aud}$/$P^{s/f}_{non\text{-}aud}$ &
Mann-Whitney U test Appendix~\ref{sec:MW-test} &
Whether audio-active-segment/non-audio-segment HII differs between success and fail runs. \\

$P^{\mathrm{sf}}_{\mathrm{JSD},\,\mathrm{A\ vs\ NA},\,S/F}$ &
Paired sign-flip permutation Appendix~\ref{sec:psf} &
Whether density between audio and non-audio segments changes significantly within successful/failed runs using $JS$ distance. \\

$P^{\mathrm{Mantel}}_{\mathrm{box\leftrightarrow path}}$ &
Mantel permutation test Appendix~\ref{sec:mantel} &
Whether run-to-run box-position distance is associated with run-to-run path-distribution distance. \\

\bottomrule
\end{tabularx}
\end{table*}

This section investigates how auditory clues such as password emitted by recorder would affect the trajectory of models. In this analysis, the trajectory is processed using the same steps in Section~\ref{sec:ana_path}. This normalization makes trajectories comparable across levels with different settings and runs under a shared coordinate frame.

The concentration of exploration is quantified by the normalized Herfindahl-Hirschman Index (HII), adapted from the Herfindahl concentration index in economics. We devide the room into \(B\times B\) occupancy grid,  and let \(p_c\) denote the probability mass of points on trajectories in cell \(c\), with \(\sum_c p_c=1\) and \(C=B^2\). The raw index is \(\mathrm{HII}_{\mathrm{raw}}=\sum_{c=1}^{C} p_c^2\), and the normalized form is
\(
\mathrm{HII}_{\mathrm{norm}}=\frac{\mathrm{HII}_{\mathrm{raw}}-1/C}{1-1/C} \in [0,1].
\)
A larger \(\mathrm{HII}_{\mathrm{norm}}\) indicates that trajectory occupancy is concentrated in fewer regions, while a smaller value indicates more spatially dispersed exploration. A detailed description for HII can be found in Appendix~\ref{app:foundations}.

Two complementary hypothesis tests are designed to separate \emph{within-run auditory effects} from \emph{between-outcome differences}.
(i) \(A\) vs \(NA\) (paired, within run): each run provides two matched observations, the audio-active segment and the non-audio segment.  
The test target is whether the typical within-run difference
\(
(\Delta_{\text{run}}=\mathrm{HII}_{\text{audio}}-\mathrm{HII}_{\text{non-audio}})
\)
is centered at zero.  
A paired two-sided Wilcoxon signed-rank test is used because the two observations are dependent and come from the same trajectory context. This design controls run-level heterogeneity and directly tests the effect of auditory condition on concentration.
(ii) \(S\) vs \(F\) (unpaired, between runs): success and fail runs are independent groups, so run-level HII distributions are compared with a two-sided Mann-Whitney \(U\) test.  

The test is performed under three segment definitions (full steps, steps with audio playing, steps with no audio) to identify where outcome separation appears. This setup tests whether exploration concentration differs by final outcome, and whether that difference is specifically amplified during audio-active periods rather than a global trajectory difference.

Results from Table~\ref{tab:hii_results2} show that \textbf{audio-active segments have significantly higher HII than non-audio segments across outcome classes, indicating more concentrated exploration with auditory cues}, indicating more concentrated exploration under auditory clues (Difficulty-2: \(p=0.0004^*\) for success and \(p=0.0215^*\) for fail; Difficulty-3: \(p=0.0039^*\) for success and \(p=0.0028^*\) for fail). The detailed run-level scatter-and-paired-line distributions are provided in Figure~\ref{fig:hii} and can further prove the consistency differences in path selection strategies.
For \(S\) vs \(F\), a significant gap appears only in audio segments of Difficulty-2 (\(p=0.0088^*\)), with success runs more concentrated than fail runs.However, difficulty-3 shows no significant \(S\) vs \(F\) gap in audio segments (\(p=0.4001\)). A plausible explanation is the additional visual key object in Difficulty-3, which dilutes isolated auditory contrast.

To verify this interpretation, a dedicated Difficulty-3 box-position analysis is conducted with a Mantel permutation test. This test is to answer if box location changes across runs, does agent path pattern change with it. The analysis compares path differences and box-location differences across runs, and also checks whether path centroids move with box coordinates. For each run, full-path distribution distance is computed by Jensen-Shannon distance between run histograms, and box-position distance is computed by Euclidean distance between aligned box coordinates.  Additional robustness tests evaluate whether the audio-related shift is stable rather than random noise. The result is significant (\(r=0.1548,\,p=0.0047^*\)), showing that visual clue position is associated with path variation and affects Difficulty-3 auditory comparisons. This evidence supports strong cross-modal coupling in EscapeCraft4D.

Further, to confirm that \textbf{audio cues systematically affect behavior rather than reflecting random variation}, trajectory-level density distributions were analyzed with permutation tests. Jensen-Shannon distance and $L_1$ effect size were used, and paired audio vs non-audio comparisons applied sign-flip permutations on within-run density differences. Validation against thousands of randomized permutations shows that observed shifts are highly unlikely by chance, with $P_{A \text{ vs } NA}^s$ and $P_{A \text{ vs } NA}^f$ reaching the permutation lower bound $0.0003^*$, demonstrating a strong and robust auditory impact on exploration strategies. All the tests we have used for analysis are listed in Table~\ref{tab:testwise_mapping_ch5_style}. For a detailed account of the statistical tests, see Appendix~\ref{app:test}.

%% file: sections/6.conclusion.tex
\section{Conclusion}
We presented \textbf{EscapeCraft-4D}, a novel 4D multimodal environment for evaluating time-aware reasoning and active cross-modal perception in Omni models and agents. Our experiments across 66 scenes of different difficulties reveal that even state-of-the-art models struggle with modality bias, transient cues, and misleading information, highlighting substantial gaps in real-world multimodal reasoning. Incorporating auditory signals and temporal constraints significantly improves navigation efficiency, cue utilization, and multi-hop reasoning, as measured by TCSS, AMR, and HII metrics. Effective modality selection and time-sensitive decision-making remain challenging, which reveal the gap between the performance on static multimodal benchmarks and emobodied, time-variant and cross-modal decision making. Overall, EscapeCraft-4D provides a comprehensive platform to advance Omni models toward realistic, dynamic, and temporally-aware multimodal intelligence, supporting future research on embodied reasoning and proactive perception in 4D environments.

%% file: sections/7.appendix.tex
\section{Evaluation Metrics}
\label{app:metric}
Follow~\cite{wang2025multimodal}, we also use average escape rate (ER) as the indicator of game completion, and five more metrics for measuring intermediate interactions, including prop gain, average steps, grab count, grab success rate, and grab ratio. Denoting the total steps as $S$, amount of succeeded grabbing action as $N_{grab}^{TP}$, grab success rate as $\text{GSR}$, grab ratio as $\text{R}_{grab}$ we have,
\begin{equation}
        \text{Prop Gain} = \frac{N_{grab}^{TP}}{\sum \text{Prop count}},  
\end{equation}
\begin{equation}
        \text{GSR} = \frac{N_{grab}^{TP}}{\sum \text{Grabbing actions}}, 
\end{equation}
\begin{equation}
        \text{R}_{grab} = \frac{\sum \text{Grabbing actions}}{S}, 
\end{equation}

We also follow~\cite{wang2025multimodal} and use the metric \textit{Intent-outcome Consistence}, denoted by $C_{IO}$, ranging [0,1] to evaluate how often the successful interaction was triggered by accident rather than the origin intention of the model. For example, a model aims at trying to open the refrigerator but opening a box instead. We use GPT-4o for auto-evaluation, and the prompt template was shown in Appendix~\ref{app:consis_prompt}. We calculated the $C_{IO}$ for Difficulty-3 across GPT-4o Omni-Agent, Gemini3-Pro-Preview and Qwen3-Omni-Thinking. Results in Table~\ref{tab:consis} implies that many completed sub-goals are triggered by accident, where Gemini3 achieves a best $C_{IO}$ with 53.75\%, while GPT-4o achieves only 37.08\% despite a best GSR of 8.39\%, indicating the gap between reasoning and perception abilities.

\begin{table}[h]
    \centering
    \caption{Consistency compared with GSR on Difficulty-3.}
    \resizebox{\columnwidth}{!}{
    \begin{tabular}{c|ccc}
    \toprule
         Models & GPT-4o Omni Agent & Gemini3-Pro-Preview & Qwen3-Omni-Thinking \\
         \midrule
         Scores: $C_{IO} (GSR)$ & 37.08 (8.39) & 53.75 (7.82) & 27.55 (0.00) \\
    \bottomrule
    \end{tabular}
    }
    \label{tab:consis}
\end{table}



\section{Measurement Foundations}
\label{app:foundations}

\subsection{Discrete Fréchet Distance}
To quantify the global structural similarity between the traversed trajectory $P$ of the agent and the optimal reference ``beeline'' path $Q$ (which is exactly the line connecting the start point of the agent and the coordinate where the exit is located), we employ the discrete Fréchet distance ($d_F$). Unlike the standard Euclidean distance between endpoints, $d_F$ accounts for the specific sequencing and spatial flow of the movement. Formally, for two polygonal curves $P = (p_1, \dots, p_n)$ and $Q = (q_1, \dots, q_m)$, the distance is defined as:
\begin{equation}
d_F(P, Q) = \inf_{\alpha, \beta} \max_{t \in [0,1]} \| P(\alpha(t)) - Q(\beta(t)) \|
\end{equation}
where $\alpha$ and $\beta$ are continuous, non-decreasing reparameterizations. A lower $d_F$ value indicates that the agent’s navigation strategy closely adheres to the geometric optimum. Detailed description for discrete Fréchet can be found in~\cite{eiter1994computing}.

\subsection{HHI-based Occupancy Index (HII)}
We measure the spatial focus of the navigation of the agent using a normalized Herfindahl-Hirschman Index (HHI) applied to a discretized occupancy grid. The environment is partitioned into $N$ equal-sized cells, where $p_i$ represents the proportion of total episode time the agent occupies cell $i$. The raw HII is defined as:
\begin{equation}
HII_{raw} = \sum_{i=1}^{N} p_i^2
\end{equation}
To ensure comparability across different room scales, we normalize the index which is broadly used in calculating hii with in the grid range such that $HII \in [0, 1]$, which is defined as:
\begin{equation}
\mathrm{HII}_{\mathrm{norm}}=\frac{\mathrm{HII}_{\mathrm{raw}}-1/C}{1-1/C} \in [0,1].
\end{equation}
We interpret a higher $HII$ as the evidence of signifying that the agent has developed a consistent, low-entropy navigation policy. Conversely, a lower $HII$ indicates a high-entropy search pattern, often seen in agents lacking sufficient sensory guidance. However, it is also possible that the model is actually stuck in a certain area, causing an abnormally high $HII$. Therefore, we subsequently used various hypothesis tests to rule out the influence of extreme cases on the analysis results. Detailed description for HII index can be found in~\cite{1964THE}.

\subsection{Path Efficiency}
Path efficiency ($\eta_{path}$) assesses the economy of movement by normalizing the theoretical shortest navigable distance $L_{min}$ by the actual distance $L_{actual}$ traversed by the agent:
\begin{equation}
\eta_{path} = \frac{L_{min}}{L_{actual}}
\end{equation}
A value of $1.0$ represents a strictly optimal trajectory. This metric isolates the ability of the agent to translate sensory inputs into direct kinetic progress. 

\subsection{Monotonic Progress Ratio}
The Monotonic Progress Ratio ($R_{mono}$) evaluates the degree of hesitation or backtracking during an episode. Let $D(t)$ be the Euclidean distance to the exit at timestep $t$. We define the progress increment as $\Delta D_t = D(t-1) - D(t)$. The ratio is the proportion of the episode duration $T$ where the agent successfully reduces the distance to the goal:
\begin{equation}
R_{mono} = \frac{1}{T} \sum_{t=1}^{T} \mathbb{I}(\Delta D_t > 0)
\end{equation}
where $\mathbb{I}$ is the indicator function. A high $R_{mono}$ indicates a persistent, goal-oriented heading.

\subsection{Survival Analysis of Exit Proximity}
To evaluate the temporal efficiency of navigation, we model the probability of an agent not reaching the exit zone using a survival analysis framework. An agent is considered to have survived (failed to escape) at normalized time $t$ if its distance to the exit remains above a critical threshold $\tau$ (e.g., $1.5m$). The survival function $S(t)$ is given by:
\begin{equation}
S(t) = P(D(t) > \tau)
\end{equation}
By analyzing the decay rate of $S(t)$ and the resulting Area Under the Curve (AUC), we can statistically determine—via Wilcoxon signed-rank tests—whether auditory guidance significantly accelerates the arrival of the agent at the exit zone compared to baseline conditions.

\subsection{Calculation of distance between trajectories}
\paragraph{Jensen-Shannon Distance} To quantify the dissimilarity between two trajectories $P$ and $Q$ of arbitrary lengths, we first discretize the room into a grid of $M \times M$ cells. Each trajectory is transformed into a discrete probability distribution, $\mathcal{P}$ and $\mathcal{Q}$, where the value of each cell represents the normalized residence steps of the agent within that spatial bin.

The divergence between these distributions is measured using the Jensen-Shannon (JS) divergence, which is a symmetrized and smoothed version of the Kullback-Leibler (KL) divergence. It is defined as:
\begin{equation}
D_{JS}(\mathcal{P} \| \mathcal{Q}) = \frac{1}{2} D_{KL}(\mathcal{P} \| \mathcal{M}) + \frac{1}{2} D_{KL}(\mathcal{Q} \| \mathcal{M})
\end{equation}
where $\mathcal{M} = \frac{1}{2}(\mathcal{P} + \mathcal{Q})$ is the average distribution. The $D_{KL}$ term represents the relative entropy:
\begin{equation}
D_{KL}(\mathcal{P} \| \mathcal{M}) = \sum_{i=1}^{M^2} \mathcal{P}_i \log \left( \frac{\mathcal{P}_i}{\mathcal{M}_i} \right)
\end{equation}
In our analysis, we utilize the \textbf{Jensen-Shannon distance}, defined as $d_{JS} = \sqrt{D_{JS}}$, which satisfies the triangle inequality. Unlike simple Euclidean distance among discrete points, $d_{JS}$ captures the complete topological difference between paths, \textbf{making it highly sensitive to detours, wall-following behaviors, and local search patterns}.

\paragraph{L1 Distance} Use the same definition of discretizing the room into a grid of $M \times M$ cells, and denote $ p_c $ and $ q_c $ as the probability the trajectory of the agent falls into each cell. The $L_1$ distance is then calculated as follows:
\begin{equation}
    L_1(\mathcal{P}, \mathcal{Q})=\sum_c{|p_c-q_c|}
\end{equation}

\section{Hypothesis tests}
\label{app:test}
We performed a series of hypothesis tests to examine differences across various conditions (e.g., success vs. failure, audio vs. non-audio) and metrics (e.g., $Hii_{norm}$, path density). The following sections provide a detailed description of the specific tests employed. A comprehensive mapping of the hypothesis tests is presented in Table~\ref{tab:testwise_mapping_ch5_style}.

\subsection{Mann-Whitney U Test}
\label{sec:MW-test}
The Mann-Whitney U test (also known as the Wilcoxon rank-sum test) is a non-parametric statistical method used to determine whether there is a significant difference between the distributions of two independent groups. Unlike the $t$-test, it does not assume a normal distribution of the data. The calculation involves the following steps:
\begin{enumerate}
    \item \textbf{Ranking:} Combine all observations from both groups and rank them from smallest to largest, assigning average ranks in the case of ties.
    \item \textbf{Test Statistic ($U$):} The $U$ statistic is calculated based on the sum of ranks for each group. For a group with size $n_1$ and rank sum $R_1$, the statistic is defined as:
    \begin{equation}
    U_1 = n_1 n_2 + \frac{n_1(n_1 + 1)}{2} - R_1
    \end{equation}
    \item \textbf{Evaluation:} The smaller of the $U$ values from the two groups is compared against a critical value table or used to calculate a $p$-value. A small $p$-value suggests that one distribution is stochastically greater than the other.
\end{enumerate}

\subsection{Wilcoxon Signed-Rank Test}
\label{sec:wilcoxon}
The Wilcoxon signed-rank test is a non-parametric alternative to the paired $t$-test, designed to compare two related samples or repeated measurements on a single sample to assess whether their population mean ranks differ. The procedure for the test is as follows:
\begin{enumerate}
    \item \textbf{Difference Calculation:} For each pair $i$, calculate the difference $d_i = x_{1,i} - x_{2,i}$. Pairs with $d_i = 0$ are typically excluded based on the ``Wilcox'' zero-handling method.
    \item \textbf{Ranking of Absolute Differences:} Rank the absolute values $|d_i|$ from smallest to largest.
    \item \textbf{Signed Ranks:} Attach the sign of the original difference to each rank. Let $W^{+}$ be the sum of ranks where $d_i > 0$, and $W^{-}$ be the sum of ranks where $d_i < 0$.
    \item \textbf{Test Statistic ($W$):} The test statistic $W$ is the smaller of the two sums calculated before: $W = \min(W^{+}, W^{-})$. The $p$-value is then derived from the distribution of $W$, indicating whether the median of the differences significantly deviates from zero.
\end{enumerate}

\subsection{Permutation-based Resampling Test}
\label{sec:perm}
When standard rank-based tests are unavailable, a permutation-based resampling approach is employed as a robust estimation of the $p$-value. This method does not rely on any specific distribution parameters. The process involves:
\begin{enumerate}
    \item \textbf{Observed Statistic:} Calculate the observed difference in means $\Delta \mu_{obs} = | \bar{x}_a - \bar{x}_b |$.
    \item \textbf{Shuffling:} Under the null hypothesis that the two groups are identical, the group labels are randomly permuted $N=5000$ times. For each permutation, a synthetic difference $\Delta \mu_{perm}$ is computed.
    \item \textbf{Probability Estimation:} The $p$-value is estimated as the proportion of permutations where $\Delta \mu_{perm} \ge \Delta \mu_{obs}$:
    \begin{equation}
    p = \frac{\sum \mathbb{I}(\Delta \mu_{perm} \geq \Delta \mu_{obs}) + 1}{N + 1}
    \end{equation}
    where $\mathbb{I}$ is the indicator function. The ``$+1$'' serves as a pseudo-count to prevent a $p$-value of zero.
\end{enumerate}

\subsection{Group-level Permutation Test}
\label{sec:group-perm}
To determine if the observed spatial distribution difference between two independent cohorts (e.g., successful vs. failed runs) is statistically significant, we perform a non-parametric permutation test. Let $\bar{\mathcal{P}}_{A}$ and $\bar{\mathcal{P}}_{B}$ be the mean occupancy probability maps of the two groups. The observed test statistic is defined as the JS distance between these mean distributions: $S_{obs} = d_{JS}(\bar{\mathcal{P}}_{A}, \bar{\mathcal{P}}_{B})$.

Under the null hypothesis that group membership has no effect on spatial behavior, we pool all runs and randomly redistribute them into two surrogate groups of sizes $n_A$ and $n_B$. This process is repeated for $N=3000$ iterations to construct a null distribution of the test statistic $S_{perm}$. The $p$-value is calculated as the proportion of permutations where $S_{perm} \geq S_{obs}$:
\begin{equation}
p = \frac{\sum_{i=1}^{N} \mathbb{I}(S_{perm, i} \geq S_{obs}) + 1}{N + 1}
\end{equation}
where $\mathbb{I}(\cdot)$ is the indicator function. This way to construct the test statistic ensures that the significance is derived from the collective spatial pattern of the group rather than being biased by individual outlier trajectories.

\subsection{Paired Sign-flip Test for Stimulus Response}
\label{sec:psf}
To isolate the direct impact of acoustic stimuli while controlling for individual agent variance, we employ a paired sign-flip test. For each run $r$, we compute the paired difference in occupancy: $\Delta_r = \mathcal{P}_{audio,r} - \mathcal{P}_{non\_audio,r}$. This represents the spatial shift caused by the audio signal within run.

The null hypothesis states that the audio trigger has no systematic effect on the path of the agent, implying that the direction of the shift $\Delta_r$ is purely stochastic. To test this, we generate $N=3000$ randomized samples where each difference map is assigned a random sign $s_r \in \{+1, -1\}$:
\begin{equation}
\bar{\Delta}_{perm} = \frac{1}{R} \sum_{r=1}^{R} s_r \cdot \Delta_r
\end{equation}
The significance is determined by comparing the observed aggregate JS distance between the actual audio and non-audio conditions against the distribution of distances generated under these random sign assignments. This paired approach is highly sensitive to subtle, consistent behavioral transitions that might be masked in a standard group-wise comparison.

\subsection{Mantel Test}
\label{sec:mantel}
The Mantel test is a statistical procedure used to evaluate the correlation between two symmetric distance (or dissimilarity) matrices, $D_X$ and $D_Y$, derived from the same set of $n$ samples. In the context of spatial navigation, this test determines whether the dissimilarity in agent trajectories correlates with the dissimilarity in environmental configurations (e.g., target prop positions). The procedure is defined as follows:

\begin{enumerate}
    \item \textbf{Matrix Construction:} Two matrices are constructed such that $D_{X(i,j)}$ represents the distance between the $i$-th and $j$-th agent trajectories (i.e. Jensen-Shannon distance of occupancy probabilities), and $D_{Y(i,j)}$ represents the Euclidean distance between the corresponding target object positions for those runs.
    \item \textbf{Observed Statistic ($r_M$):} The Mantel correlation coefficient is computed using the standardized Pearson or Spearman correlation between the corresponding off-diagonal elements of the two matrices:
    \begin{equation}
    r_M = \frac{1}{m-1} \sum_{i=1}^{n-1} \sum_{j=i+1}^{n} \text{std}(D_{X(i,j)}) \cdot \text{std}(D_{Y(i,j)})
    \end{equation}
    where $m = n(n-1)/2$ is the number of unique pairs.
    \item \textbf{Permutation Procedure:} Since elements within distance matrices are not independent, a permutation-based approach is required for significance testing. The rows and columns of one matrix (e.g., $D_X$) are randomly shuffled $N=3000$ times while keeping the other matrix $D_Y$ fixed. For each shuffle $k$, a permuted statistic $r_{M,k}$ is calculated.
    \item \textbf{Significance Evaluation:} The $p$-value is defined as the proportion of permutations where the absolute value of the permuted statistic equals or exceeds the observed statistic:
    \begin{equation}
    p = \frac{\sum \mathbb{I}(|r_{M,k}| \geq |r_{M,obs}|) + 1}{N + 1}
    \end{equation}
    A significant $p$-value indicates that the spatial distribution of the trajectories is non-randomly associated with the positioning of environmental landmarks.
\end{enumerate}

\section{Prompt Template}\label{app:prompt}
We follow~\cite{wang2025multimodal} and define the following prompts.

\paragraph{System Prompt} Compare to~\cite{wang2025multimodal}, we added one more action space for auditory modality to the system prompt. The complete prompt is shown in table~\ref{tab:system_prompt}.

\begin{table*}[h]
\centering
\caption{The System Prompt} 
\label{tab:system_prompt} 
\begin{tabularx}{\textwidth}{Y}
\toprule
   \rowcolor{gray!10} \multicolumn{1}{c}{\textit{Instruction Prompt}} \\
You find yourself locked inside a room, and your ultimate goal is to escape the room. i.e. the room escape game.\\\\
You can explore the room, interact with objects, inspect items, and resolve puzzles. You can adopt the following actions to explore the room and interact with objects:\\
    \midrule
   \rowcolor{gray!10} \multicolumn{1}{c}{\textit{Operation Prompt}} \\
   - move\_forward: float, ranged between [-10, 10]. This is the number of meters you want to move forward (negative value means moving backward). \\
- rotate\_right: float, ranged between [-180, 180]. This is the number of degrees you want to turn right (negative value means turn left). \\
- rotate\_down: float, ranged between [-90, 90]. This is the angle you want to adjust your view vertically. Positive value means looking downward, while a negative value means looking upward. Angle 0 means looking straight ahead. \\
- jump: bool, whether you want to jump (can be used together with moving forward), e.g., True represents the action ``to jump''. \\
- look\_at: list[x: foat, y: float], the range of x and y is [0, 1]. This parameter is the coordinates of the point in the image you want to look at. For reference, the coordinates of the upper left corner of the scene are (0, 0) and the coordinates of the lower right corner are (1, 1). Also to mention that there are on clues on the ceiling. \\
- grab: bool, whether you require to interact with the object located exactly at the center of the scene (marked by a red dot). e.g., to grab the key or to interact with (or open) a box at the center of the scene, set grab=True. The red dot assists in locating the object you require to interact with. You might need to adjust the view or move closer to ensure the red dot is on your target object, through the rotate\_right, rotate\_down, and move\_forward actions. To successfully grab an object, you should center the object via the red dot and be in a certain distance to it. If the grabbing fails, try move closer towards the object. If it fails multiple times at the same position, you should be aware that not all objects are interactable, do not get stucked in uninteractable position. \\
- trigger: bool, whether you want to attempt to trigger the sound of a sound-emitting object (e.g., recorder/radiogram/musicbox) at the center of the scene. This ONLY works when you are close enough and facing it; otherwise it will fail and you should move closer / adjust your view. \\
- interactions : dict:{``use\_item\_id'': str, this is the item\_id you require to view or use (when used together with grab=True, it means to use this item to interact with the target object you want to grab, e.g. using item\_id of the key to open the door in the scene), ``input'': str, this is the message you want to input when interacting with the center object}. \\
- read: str, this is the item\_id that you want to get detailed information from your bag. \\
- rationale: str, represents the rationale of your action. This should explain your decision-making process and help the agent understand your thinking process.\\\\

You need to return data in the following format of JSON\_string to interact with the scene:\\
\texttt{\{}\\
\hspace{2em}\texttt{       ``move\_forward'': float,}\\
\hspace{2em}\texttt{       ``rotate\_right'': float,}\\
\hspace{2em}\texttt{       ``rotate\_down'': float,}\\
\hspace{2em}\texttt{       ``jump'': bool,}\\
\hspace{2em}\texttt{       ``look\_at'': [x: float, y: float],}\\
\hspace{2em}\texttt{       ``grab'': bool,}\\
\hspace{2em}\texttt{       ``trigger'': bool,}\\
\hspace{2em}\texttt{       ``interactions'': \{}\\
\hspace{4em}\texttt{           ``use\_item\_id'': str,}\\
\hspace{4em}\texttt{           ``input'': str}\\
\hspace{2em}\texttt{       \},}\\
\hspace{2em}\texttt{       ``read'': str,}\\
\hspace{2em}\texttt{       ``rationale'': str}\\
\texttt{   \}}\\\\

All of the above operations are optional. If no value is passed in, the interactive operation will not be performed.\\\\

You must follow the above instructions and don't say anything else except for the JSON\_string of operations.\\
\bottomrule
\end{tabularx}
\end{table*}

\paragraph{Step Prompt} The complete prompt is shown in table~\ref{tab:step_prompt}.

\begin{table}[H]  
\centering
    \caption{The Step Prompt} 
    \label{tab:step_prompt} 
    \begin{tabularx}{\linewidth}{Y}
    \toprule
   \rowcolor{gray!10} \multicolumn{1}{c}{\textit{Interaction Result}} \\
\{interaction\_result\} \\
===\\
    \midrule
   \rowcolor{gray!10} \multicolumn{1}{c}{\textit{Inventory}} \\
   
The items in your bag usable include:\\
\{bag\_desc\}\\
===\\
\midrule
   \rowcolor{gray!10} \multicolumn{1}{c}{\textit{Step Prompt}} \\
   Please determine the next action(s) that could help you observe the room or obtain useful tools or clues.\\
If you find yourself stuck in a corner, try turn around by passing rotate\_right.\\
You need to return data in the following format of JSON\_string to interact with the scene and don't say anything else: \\
\texttt{\{}\\
\hspace{2em}\texttt{       ``move\_forward'': float,}\\
\hspace{2em}\texttt{       ``rotate\_right'': float,}\\
\hspace{2em}\texttt{       ``rotate\_down'': float,}\\
\hspace{2em}\texttt{       ``jump'': bool,}\\
\hspace{2em}\texttt{       ``look\_at'': [x: float, y: float],}\\
\hspace{2em}\texttt{       ``grab'': bool,}\\
\hspace{2em}\texttt{       ``trigger'': bool,}\\
\hspace{2em}\texttt{       ``interactions'': \{}\\
\hspace{4em}\texttt{           ``use\_item\_id'': str,}\\
\hspace{4em}\texttt{           ``input'': str}\\
\hspace{2em}\texttt{       \},}\\
\hspace{2em}\texttt{       ``read'': str,}\\
\hspace{2em}\texttt{       ``rationale'': str}\\
\texttt{   \}}\\
    \bottomrule
\end{tabularx}
\end{table}

\paragraph{Prompt for Consistency Evaluation}\label{app:consis_prompt} Compared to~\cite{wang2025multimodal}, we found that prevailing models seem to prefer using more rationale to describe how to adjust the perspective, thus interfering with the judgement of the scoring model. Therefore, we added an additional special case to accommodate this situation. The complete prompt is shown in table~\ref{tab:consistency_prompt}.

\begin{table*}[htb]
\centering
    \caption{The Consistency Evaluation Prompt}
    \label{tab:consistency_prompt}
    \begin{tabularx}{\textwidth}{Y}
    \toprule
    \rowcolor{gray!10} \multicolumn{1}{c}{\textit{Consistency Evaluation Prompt}} \\
You are a reasoning consistency evaluator for a multimodal agent benchmark. Your task is to determine whether the agent's intention (described in the ``rationale'') matches the actual interaction result (described in the ``response'').\\\\

Specifically, you are given: \\
- The agent's \textbf{rationale} for why it attempted an interaction, describing its goal or belief about the current environment.\\
- The \textbf{response} from the environment after the interaction, which includes the actual result (e.g., what item was interacted with and what was obtained).\\\\

Your goal is to determine whether the object the agent intended to interact with \textbf{matches} the object that was actually interacted with according to the response.\\\\

If the agent tried to interact with object A (e.g., a microwave), but the response shows interaction with object B (e.g., a box), and B was not the intended target, this is considered an \textbf{inconsistent interaction} (i.e., accidental success or misaligned action).\\\\

\textbf{Special Cases}: if the response is “Escaped successfully!”, you should check whether the agent's rationale explicitly or implicitly indicates the intention to escape (e.g., trying to open the door to leave). In another special case, if the agent is trying to adjust its view to align with the door, and it has triggered the information of the door, this is considered consistent. If not, treat it as inconsistent.\\\\

You must give your feedback in the following JSON-string format and \textbf{DON'T} say anything else: \\
\texttt{\{} \\
\hspace{2em} \texttt{``Consistency'': 1 | 0} \\
\texttt{\}} \\\\

Where: \\
- \texttt{1} means the rationale and interaction are consistent (i.e., aligned).\\
- \texttt{0} means the interaction appears to be accidental, mismatched, or unintended.\\\\

Respond \textbf{only} with \texttt{Consistency: 1} or \texttt{Consistency: 0}.\\\\

\textbf{---}\\\\

\textbf{Example 1:} \\
rationale: \emph{Moving closer to the microwave to try interacting with it one last time using '1264'. If this doesn't work, I'll need to explore other areas.}\\
response: \emph{You used the correct password to unlock the box... You did not interact with any objects in the last step.}\\
\textbf{Expected output:} \texttt{Consistency: 0} \\\\

\textbf{Example 2:} \\
rationale: \emph{I’ll try using the key I just picked up on the door. Let’s see if I can escape now.}\\
response: \emph{Escaped successfully!}\\
\textbf{Expected output:} \texttt{Consistency: 1} \\\\

\textbf{Example 3:} \\
rationale: \emph{I want to check if the small box has any useful items inside.}\\
response: \emph{You opened the box and found a screwdriver.}\\
\textbf{Expected output:} \texttt{Consistency: 1} \\\\

\textbf{Please score the following interaction:} \\
rationale: \{\texttt{rationale}\} \\
response(s): \{\texttt{response}\} \\
    \bottomrule
\end{tabularx}
\end{table*}